\documentclass[10pt
  ]{article}

\usepackage[top=0.9in, bottom=0.9in, left=1.0in, right=1.0in]{geometry}

\usepackage{wrapfig}
\usepackage[round]{natbib}

\usepackage{hyperref}

\usepackage{amsmath}
\usepackage{amssymb}
\usepackage{adjustbox}
\usepackage{mathtools}
\usepackage{amsthm}

\usepackage{caption}

\theoremstyle{plain}

\theoremstyle{definition}

\theoremstyle{remark}

\usepackage[utf8]{inputenc} %
\usepackage[T1]{fontenc}    %
\usepackage{url}            %
\usepackage{booktabs}       %
\usepackage{amsfonts}       %
\usepackage{nicefrac}       %
\usepackage{xcolor}         %

\usepackage{amsmath,amsfonts,bm,amsthm,amssymb,mathrsfs,algorithm,algorithmic}
\usepackage{hyperref,url,booktabs,enumitem,multicol,graphicx,caption,subcaption}

\usepackage{adjustbox}
\usepackage{hhline}
\usepackage{multirow}
\usepackage{arydshln}

\def\eqref#1{(\ref{#1})}

\def\1{\bm{1}}

\newcommand{\bE}{\mathbb{E}}

\usepackage{stackengine}

\makeatletter
\newcommand{\distas}[1]{\mathbin{\overset{#1}{\kern\z@\sim}}}%

\newcommand{\beqs}{\vspace{0mm}\begin{eqnarray}}
\newcommand{\eeqs}{\vspace{0mm}\end{eqnarray}}
\newcommand{\barr}{\begin{array}}
\newcommand{\earr}{\end{array}}

\newcommand{\vv}{\boldsymbol{v}}

\newcommand{\xv}{\boldsymbol{x}}

\newcommand{\zv}{\boldsymbol{z}}

\newcommand{\epsilonv}{\boldsymbol{\epsilon}}

\newcommand{\Ls}{\mathcal{L}}

\newcommand{\given}{\,|\,}

\usepackage{amsmath}
\usepackage{amssymb}
\usepackage{mathtools}
\usepackage{amsthm}
\usepackage{graphicx}
\usepackage{cleveref}
\usepackage{multirow}

\usepackage{tabularray}
\UseTblrLibrary{booktabs}

\usepackage{color, colortbl}
\definecolor{hl_color}{gray}{0.925}

\newcommand{\proteina}{Prote{\'i}na }

\usepackage{bbm}
\usepackage{algorithm}
\usepackage{algorithmic}

\usepackage{authblk}

\usepackage{geometry}

\usepackage[final,expansion=alltext]{microtype}
\usepackage[english]{babel}
\usepackage[parfill]{parskip}
\usepackage{afterpage}
\usepackage{framed}

\renewenvironment{abstract}%
{%
  \vskip 0.075in%
  \centerline%
  {\large\bf Abstract}%
  \vspace{0.5ex}%
  \begin{quote}%
}
{
  \par%
  \end{quote}%
  \vskip 1ex%
}

\usepackage{longtable}
\usepackage{mdframed}

\usepackage{listings}
\definecolor{codegreen}{rgb}{0,0.6,0}
\definecolor{codegray}{rgb}{0.5,0.5,0.5}
\definecolor{codepurple}{rgb}{0.58,0,0.82}
\definecolor{backcolour}{rgb}{0.95,0.95,0.92}

\lstdefinestyle{mystyle}{
    backgroundcolor=\color{backcolour},   
    commentstyle=\color{codegreen},
    keywordstyle=\color{magenta},
    numberstyle=\tiny\color{codegray},
    stringstyle=\color{codepurple},
    basicstyle=\ttfamily\footnotesize,
    breakatwhitespace=false,         
    breaklines=true,                 
    captionpos=b,                    
    keepspaces=true,                 
    numbers=left,                    
    numbersep=5pt,                  
    showspaces=false,                
    showstringspaces=false,
    showtabs=false,                  
    tabsize=2
}

\title{SPID: Distilled Protein Backbone Generation}

\author[1]{Liyang Xie}
\author[1]{Haoran Zhang}
\author[1]{Zhendong Wang}
\author[2]{Wesley Tansey}
\author[1]{Mingyuan Zhou}
\affil[1]{The University of Texas at Austin}
\affil[2]{The Memorial Sloan Kettering Cancer Center}
\date{}

\begin{document}
\maketitle

\begin{abstract}
Diffusion- and flow-based generative models have recently demonstrated strong performance in protein backbone generation tasks, offering unprecedented capabilities for \textit{de novo} protein design. However, despite their generation quality, these models are constrained by slow sampling, often requiring hundreds of iterative steps. This computational bottleneck limits their practical utility in large-scale protein discovery, where thousands to millions of candidate structures are needed. To address this challenge, we explore the techniques of score distillation, which has shown great success in reducing the number of sampling steps in the vision domain while maintaining high generation quality. However, a straightforward adaptation of these methods results in unacceptably low designability. We introduce Score Protein identity Distillation (SPID), which resolves this incompatibility by combining few-step generation with inference-time noise scaling. SPID adapts the Score identity Distillation (SiD) framework to both diffusion- and flow-based models without requiring access to pretraining data.
Applied to the \proteina flow-matching model, our 16-step generator achieves 94.4\% designability, matching the 400-step teacher (94.2\%; $p = 0.89$ at $n = 500$), while delivering more than a 20-fold reduction in effective backbone-generation time and maintaining comparable diversity and novelty at lengths 50--256; longer chains require a separately distilled 32-step generator. SPID generalizes across unconditional generation, fold-class conditional generation, and motif scaffolding, and extends to equivariant diffusion architectures, achieving significant reduction in generation time with comparable generation quality to the teacher in all tasks.
The resulting reduction in inference cost could facilitate large-scale \textit{in silico} protein design, thereby advancing diffusion-based models toward real-world protein engineering applications.
\end{abstract}

\section{Introduction}
\textit{De novo} protein design aims to generate novel proteins with user-specified biological functions, bypassing the domain expertise and laboratory experimentation required by traditional approaches such as directed evolution \citep{Huang2016protein_design, Dougherty2009directed_evolution}.
Deep generative models have shown great success in tasks like image generation, but adapting them from images to proteins requires addressing fundamental structural constraints. Proteins have no fixed canonical orientation, motivating SE(3)-equivariant architectures. Models must respect physical constraints and track pairwise inter-residue distances. Crucially, unlike images where local errors yield merely unrealistic outputs, errors in local or global backbone geometry, such as inconsistencies in pairwise distances or torsion angles, can cascade into loss of key biochemical features and render structures completely undesignable \citep{Eguchi2022Ig_VAE, Anand2022Protein_equivariant}.
Architectural innovations from AlphaFold2 \citep{Jumper2021AF2}, including invariant point attention (IPA) and triangle attention layers, addressed the first two challenges, enabling models such as RFDiffusion \citep{watson2023RFDiffusion}, FrameDiff \citep{yim2023framediff}, and Genie \citep{lin2023genie} to generate high-quality, designable backbone structures encoding genuinely novel folds. Meanwhile, Chroma \citep{Ingraham2023Chroma} took a different architectural approach, achieving similar goals via a sub-quadratic random graph neural network. More recent work has shown that even without equivariant architecture, flow-based models such as \proteina \citep{geffner2025proteina} can achieve state-of-the-art designability.

\textbf{Sampling speed as a bottleneck.}
A critical obstacle limits the practical utility of these models: \textbf{sampling speed}.
Generating a single backbone requires hundreds or even 1000 iterative steps, with each step often involving expensive IPA and triangle computations.
This is a fundamental barrier to large-scale protein design, where biologists must screen tens of thousands of candidate structures to evaluate stability, binding affinity, or other functional properties. For example, using \proteina\!\!'s reported sampling times, with one A6000-48GB GPU, a 100{,}000 backbone-design campaign would cost around 150 GPU-hours.
Slow generation becomes a bottleneck in iterative generate--evaluate--test workflows, limiting throughput and delaying discovery.
Accelerating backbone generation is therefore not merely a matter of computational efficiency but a prerequisite for practical deployment.
Prior work has made partial progress: flow-based FrameFlow \citep{yim2023frameflow} improved sampling speed over its diffusion predecessor FrameDiff \citep{yim2023framediff}, and \proteina \citep{geffner2025proteina} demonstrated that an architecture without IPA and triangle layers can still achieve competitive designability.
Nonetheless, \proteina still requires 400 sampling steps, leaving significant room for acceleration.

\textbf{Distillation in the image domain.}
Score distillation methods have dramatically reduced sampling costs in image generation, training one- or few-step generators that match or exceed the quality of hundred-step models \citep{Fan2025survey_distillation, ma2024survey_diffusion, shen2025efficient_survey}.
Score identity Distillation (SiD) \citep{zhou2024sid} and its extensions \citep{zhou2025sida, zhou2025sidfewstep} achieve state-of-the-art one- and few-step image generation by matching the generator's score function to that of a pretrained teacher, with the objective of minimizing a Fisher divergence-based loss.
However, as we discuss next, a property unique to protein backbone generation prevents direct application of these methods.

\textbf{The noise rescaling challenge.}
A further complication specific to protein backbone generation compounds the difficulty of acceleration.
Because small geometric errors readily produce undesignable structures, protein backbone generators universally apply \emph{inference-time noise rescaling}: lowering the sampling temperature to enforce structural precision at the cost of some diversity.
While effective for standard multi-step sampling, this practice creates a fundamental mismatch for distillation: distillation methods are trained under the original noise schedule, and rescaling at inference breaks the alignment between the training and inference distributions.
As a result, naively applying an off-the-shelf distillation method leads to near-zero designability.

\textbf{This work.}
We introduce \textbf{Score Protein identity Distillation (SPID)}, a distillation framework that resolves this incompatibility through two key ingredients: few-step generation and inference-time noise scaling.
SPID adapts the SiD framework \citep{zhou2024sid} to both diffusion- and flow-based backbone generators and supports data-free training.
Our main contributions are:
\begin{itemize}
    \item \textbf{Identifying the failure mode of one-step distillation for proteins.}
    We show that one-step generators trained under score distillation achieve near-zero designability (0.4\%),
    and that alternative distillation objectives and hyperparameter tuning do not recover it. The obstacle is structural: $\gamma$ rescales the noise injected at each \emph{intermediate} step, so a one-step generator offers no injection point for it, and the one operation that is available, scaling the initial Gaussian noise, recovers nothing (Appendix~\ref{subsec:one_step}).

    \item \textbf{SPID: few-step generation with inference-time noise scaling.}
    We propose combining few-step generation with the noise rescaling practice standard in protein backbone generation.
    Noise rescaling is applied only at inference, preserving the training distribution while enabling low-temperature sampling.
    Applied to \proteina \citep{geffner2025proteina}, our 16-step generator achieves 94.4\% designability, matching the 400-step teacher model (94.2\%; $p = 0.89$), with a more than 20-fold reduction in effective backbone-generation time and maintained diversity and novelty.

    \item \textbf{Generality across tasks and architectures.}
    We demonstrate that SPID generalizes across unconditional generation, fold-class conditional generation, and motif scaffolding tasks.
    It also applies to equivariant diffusion architectures, distilling the 1000-step Genie model \citep{lin2023genie} into a 10-step generator with matching (according to scTM) or better (according to scRMSD) designability.

\end{itemize}
To our knowledge, SPID is the first successful distillation of a protein backbone generator.
By reducing sampling from hundreds of steps to as few as 16, it makes large-scale \textit{in silico} screening of thousands to millions of candidate structures tractable, enabling tighter integration of generative models with practical protein engineering workflows.

\section{Background and Related Work}\label{sec:background}
\textbf{Protein structure representation.}
A protein backbone is typically represented by the $N$-$C_\alpha$-$C$-$O$ atoms of each residue.
The most expressive representation uses rigid-body \emph{frames} centered on each $C_\alpha$ atom, capturing both the residue position and local orientation via a rotation matrix derived from the N-$C_\alpha$-C atom positions.
Because proteins have no canonical orientation, frame-based representations are typically modeled by SE(3)-equivariant networks with a diffusion process that handles both translations and rotations.
An alternative, simpler representation uses only the $C_\alpha$ coordinates, expressing the backbone as an $N\times3$ matrix of Cartesian coordinates.
Although this discards orientation information, works such as Genie \citep{lin2023genie} and \proteina \citep{geffner2025proteina} have demonstrated that models whose diffusion process operates on $C_\alpha$ coordinates can achieve strong designability, while enabling simpler, more scalable training.

\textbf{Flow matching.} We take rectified flow \citep{Liu2023rectifiedFlow} as an example, as it is the basis of our teacher model \proteina \citep{geffner2025proteina}. It defines a linear interpolant between noise and data: given real data $\xv_d$ and noise $\epsilonv \sim \mathcal{N}(0, \mathbf{I})$, the noised sample at time $t \in [0,1]$ is
\begin{equation}\label{eq:rectified_flow_interpolant}
    \xv_t = t\xv_d + (1-t)\epsilonv,
\end{equation}
and the network is trained to regress the target velocity
\begin{equation}
    \vv_t^\phi(\xv_t,t) \approx \xv_d - \epsilonv.
\end{equation}
This is equivalent to a Gaussian diffusion process with $a_t = t$ and $\sigma_t = 1-t$, and as shown in \citet{Albergo2023UnifyingFlowDiffusion}, flow matching and diffusion models with Gaussian noise are equivalent upon reparametrization.
This connection is key: the score-based distillation ideas of SiD transfer to flow matching models with appropriate adjustments, as we derive in the Method section.

\textbf{Diffusion distillation with SiD.}
In SiD \citep{zhou2024sid}, the teacher model $f_\phi$ provides a score estimate $S_\phi(\xv_t) = \sigma_t^{-2}(a_t f_\phi(\xv_t, t) - \xv_t)$, where $f_\phi(\xv_t, t) \approx \mathbb{E}(\xv_d|\xv_t)$ is the teacher's clean-data prediction.
This score is used as a proxy for the true data score $\nabla_{\xv_t} \ln p_\text{data}(\xv_t)$.
A generator $G_\theta$ is trained so that its induced distribution matches the data distribution, by training a secondary \emph{fake score network} $f_\psi$ to track the generator's score and minimizing the Fisher divergence between $S_\phi$ and $S_\psi$.
A key advantage of SiD is that it supports data-free training, as it requires only the pretrained teacher and fresh noise samples to calculate the Fisher divergence. On the other hand, when training data are available, they can be incorporated via adversarial components as in \citet{zhou2025sida}.

\textbf{Related work in protein backbone generation.}
RFDiffusion \citep{watson2023RFDiffusion} fine-tunes the RoseTTAFold structure-prediction network as a DDPM, adding Gaussian noise to translations and Brownian motion to rotations in the SE(3) frame representation.
Chroma \citep{Ingraham2023Chroma} uses a correlated polymer diffusion process with a sub-quadratic random graph neural network, enabling generation of very long proteins and complexes.
FrameDiff \citep{yim2023framediff} applies a theoretically grounded SE(3) diffusion directly over backbone frames using IPA.
FrameFlow \citep{yim2023frameflow} and FoldFlow \citep{Bose2024FoldFlow} adopt flow matching over SE(3) frames as a faster alternative.
Proteus \citep{Wang2024proteus} improves efficiency by replacing dense triangle layers with a graph-based approximation.
SMCDiff \citep{Trippe2023SMCDiff} pioneered the use of a $C_\alpha$-coordinate diffusion process for backbone generation.
Genie \citep{lin2023genie}, Genie2 \citep{lin2024genie2}, and \proteina \citep{geffner2025proteina} subsequently demonstrated that such models can match or exceed frame-based methods in designability, with the added benefit of simpler training and better scalability.

\textbf{Related work in diffusion distillation.}
Progressive distillation \citep{Salimans2022progressivedistillation} was among the first methods to reduce sampling steps, halving them at each distillation iteration.
Consistency models \citep{Song2023ConsistencyModels} learn the full ODE trajectory, enabling direct denoising from any point along it.
Consistency Trajectory Models \citep{kim2024CTM} and Flow Map Matching \citep{boffi2025flowmapmatching} extend this to model flow maps between arbitrary trajectory points, and Align-Your-Flow \citep{sabour2025AYF} demonstrates strong few-step image generation results in this family.
Distribution-matching approaches, beginning with Diffusion-GAN \citep{Wang2023diffusiongan}, instead minimize a divergence between the generator distribution $p_\theta(\xv_t)$ and the data distribution at every noise level~$t$.
\citet{wang2023vsd} and \citet{Luo2024diffinstruct} minimize the KL divergence, while SiD \citep{zhou2024sid} uses the Fisher divergence for data-free distillation driven entirely by the teacher's score function; its adversarial extension SiDA \citep{zhou2025sida} incorporates real training data to further improve quality.
SiD-FewStep \citep{zhou2025sidfewstep} extends SiD with a theoretically grounded multi-step training framework.
To our knowledge, none of these approaches have been successfully applied to protein backbone generation, a gap we attribute to the structural sensitivity of protein backbones to geometric error and the resulting incompatibility between noise rescaling and one-step generation.

\section{Method}
SPID adapts SiD \citep{zhou2024sid} to protein backbone generation through two key ingredients: few-step generation and inference-time noise scaling.
In Section~3.1 we derive the SiD distillation objective for rectified flow.
Section~3.2 extends this to few-step generators.
Section~3.3 introduces inference-time noise scaling, the ingredient that makes designability recoverable.
Section~3.4 covers additional training details.
We focus on the rectified flow formulation \citep{Liu2023rectifiedFlow} and use the $\mathcal{M}_\text{FS}^\text{no-tri}$ model, the fastest \proteina variant, as the teacher.

\subsection{SiD for Flow Matching}
As established in \citet{Albergo2023UnifyingFlowDiffusion} and \citet{geffner2025proteina}, flow matching and diffusion models are equivalent upon reparametrization and the key quantity linking the two classes of models is the score function, which is also the core of SiD \citep{zhou2024sid}.
SiD estimates the score of the data distribution as $\nabla_{\xv_t} \text{ln}~p_{\text{data}}(\xv_t) \approx S_\phi(\xv_t) = \sigma_t^{-2} (a_t f_\phi (\xv_t, t) - \xv_t)$, where $f_\phi(\xv_t, t) \approx \mathbb{E}(\xv_d|\xv_t)$ is the predicted true data $\xv_d$ by the pretrained model conditioned on the noised data $\xv_t$ at timestep $t$. For the rectified flow of Eq.~\eqref{eq:rectified_flow_interpolant} these specialize to $a_t = t$ and $\sigma_t = 1-t$, which is why $t$ appears in place of $a_t$ below. The distillation objective minimizes the score difference between the pretrained network $f_\phi$ and the generator network $G_\theta$ for all $t$:
\begin{equation} \label{eq:score_diff}
    \text{min}_\theta \mathbb{E}_{\xv_t \sim p_\theta(\xv_t)} [\|S_\phi(\xv_t) - \nabla_{\xv_t} \text{ln}~p_\theta(\xv_t)\|_2^2]
\end{equation}

In order to apply SiD to a flow matching model, we first derive the score function for rectified flow as $\nabla_{\xv_t} \text{ln}~p_{\text{data}}(\xv_t) \approx (1-t)^{-2} (t f_\phi(\xv_t,t) - \xv_t)$ where $f_\phi(\xv_t, t) = \xv_t + (1-t) v_t^\phi(\xv_t,t)$. To estimate the score of the generator in Eq.~\eqref{eq:score_diff}, we train another neural network $f_\psi$ using the usual flow matching loss. We treat the generator output $\xv_g$ as clean data and define the fake score training objective:
\begin{equation} \label{eq:fake_score_loss}
    \text{min}_\psi \mathbb{E}_{q(\xv_t|\xv_g, t) p_\theta(\xv_g)} [\frac{1}{N(1-t)^2}\|f_\psi(\xv_t, t) - \xv_g\|_2^2]
\end{equation}
where $N$ is the number of residues.

Following \citet{zhou2024sid}, we derive the initial approximation to the target score difference in Eq.~\eqref{eq:score_diff} as:
\begin{equation} \label{eq:L1}
    \begin{split}
        \Ls^{(1)}_\theta = &\mathbb{E}_{\xv_t\sim p_\theta(\xv_t)} \Big[\|\frac{t}{(1-t)^2} \delta_{\phi,\psi}(\xv_t,t)\|_2^2\Big], \\
        & \delta_{\phi,\psi}(\xv_t,t) = f_\phi(\xv_t,t) - f_\psi(\xv_t,t)
    \end{split}
\end{equation}
and the projected score matching loss as:
\begin{equation} \label{eq:L2}
    \begin{split}
        \Ls^{(2)}_\theta = &\frac{t^2}{(1-t)^4} \bE_{\xv_t \sim p_\theta(\xv_t), \xv_t \sim q(\xv_t\given \xv_g, t)} \Big[ \\ & (f_\phi(\xv_t, t) - f_\psi(\xv_t,t))^T (f_\phi(\xv_t,t) - \xv_g) \Big]
    \end{split}
\end{equation}
As discussed in \citet{zhou2024sid}, minimizing $\Ls^{(1)}_\theta$ alone is insufficient, and a fused loss of both $\Ls^{(1)}_\theta$ and $\Ls^{(2)}_\theta$ in the form of $\Ls_\theta = \Ls^{(2)}_\theta - \alpha \Ls^{(1)}_\theta$ needs to be considered. We set $\alpha=1.0$ in all experiments, consistent with the range $\alpha \in [0.75, 1.2]$ reported in \citet{zhou2024sid} (see Appendix~E).

With derivations detailed in Appendix~F, the final generator loss takes the form:
\begin{equation} \label{eq:generator_loss}
    \begin{split}
        \Ls_\theta = &\frac{\omega(t)t^2}{(1-t)^4} \Big[(1-\alpha)\|f_\phi(\xv_t,t)-f_\psi(\xv_t, t)\|_2^2 + \\
        & (f_\phi(\xv_t,t) - f_\psi(\xv_t,t))^T (f_\psi(\xv_t,t) - \xv_g) \Big]
    \end{split}
\end{equation}
And the generator loss reweighting function $\omega(t)$ is defined~as:
\begin{equation} \label{eq:omega_t}
    \omega(t) = \frac{(1-t)^4}{Nt^2\|\xv_g - f_\phi(\xv_t, t)\|_{1,\text{sg}}}
\end{equation}

We defer the detailed derivation relating the velocity $\vv_t^\phi(\xv_t,t)$ to the neural network output $f_\phi(\xv_t,t)$ to Appendix~F.

\subsection{Few-Step Generation}\label{sec:few_step}
Although one-step SiD generators perform well in image generation, our one-step protein generator produces almost no designable samples (0.4\%, Tab.~\ref{tab:main_result}).
Switching to another distillation objective such as Diff-Instruct \citep{Luo2024diffinstruct}, changing the loss reweighting, and tuning the training hyperparameters do not recover it either.
The obstacle is structural: $\gamma$ in Eq.~\eqref{eq:noise_scale_sampling} rescales the noise injected at each \emph{intermediate} step, so a generator with no intermediate steps has no injection point for it. The one operation that \emph{is} available at $K=1$ --- scaling the initial Gaussian noise --- recovers nothing: over $\gamma_0 \in \{0.2, \dots, 1.0\}$ crossed with $t_{init} \in \{0.2, 0.37, 0.6\}$, all give 0.0\% or near-0.0\% designability, for both an undistilled teacher and two distilled one-step generators (Appendix~\ref{subsec:one_step}). Raising the step count alone is not sufficient either --- without rescaling, designability reaches only 37.0\% at $K = 20$ --- so both ingredients are required. To allow low temperature sampling, SPID modifies the SiD Few-Step framework \citep{zhou2025sidfewstep} to distill few-step generators. Samples are generated by chaining $K$~steps:

\begin{equation}
\label{eq:uniform_step_matching}
    \xv_g^{(k)} = G_\theta(t_k \text{sg}(\xv_g^{(k-1)})+(1-t_k) \epsilonv_k), \; \epsilonv_k \sim \mathcal{N}(0,\mathbf{I})
\end{equation}
where $t_k$ is the $k$th step in a predefined time path $t_{steps}$, $\xv_g^{(k)}$ is the generated sample at step $k$, and sg stands for the stop gradient function, which is important during training since we use the uniform-step matching approach specified in \citet{zhou2025sidfewstep}.

During training, we uniformly sample a step $k \in \{1, ..., K\}$ and generate $\xv_g^{(k-1)}$ using Eq.~\eqref{eq:uniform_step_matching}. Only the final step from $\xv_g^{(k-1)}$ to $\xv_g^{(k)}$ is differentiated, avoiding backpropagation through the entire generation chain.

\subsection{Inference Time Noise Scaling}\label{sec:noise_scaling}

Our few-step generators distilled from the strategy above still only achieve underwhelming designability ($28.2\%$ at $K = 16$, Tab.~\ref{tab:step_gamma_scan}) using the standard sampling procedure. Now that our few-step generators take multiple steps, we can apply the same noise rescaling at each step. Critically, rescaling is applied \emph{only at inference}, leaving the training distribution unaffected. We modify the sampling step of Eq.~\eqref{eq:uniform_step_matching} as:
\begin{equation}
\label{eq:noise_scale_sampling}
    \xv_g^{(k)} = G_\theta(t_k \xv_g^{(k-1)}+\gamma(1-t_k) \epsilonv_k)
\end{equation}
 where $\gamma$ is the noise scaling factor and $\epsilonv_k \sim \mathcal{N}(0,\mathbf{I})$. The choice of the noise scale $\gamma$ greatly affects the generation quality as discussed in Appendix \ref{app:gamma_frontier}. We set $\gamma = 0.45$, which gives the optimal balance between designability and diversity in the evaluation data according to the full 500-sample $\gamma$ sweep in Appendix \ref{app:gamma_frontier}. Interestingly, the value coincides with \proteina\!\!'s published optimal $\gamma$.

Eq.~\eqref{eq:uniform_step_matching} (without noise rescaling) is used when generating samples during distillation training. On the other hand, Eq.~\eqref{eq:noise_scale_sampling} (with noise rescaling) is used at inference when producing actual protein structures for evaluation.

\subsection{Additional Training Details}
\textbf{Number of Residues.} Reweighting both the generator loss and the fake score loss by $N$, the number of residues in the protein structure, is crucial so that the generator learns to generate proteins of different lengths. For unconditional distillation, we do not require any existing databases like the PDB \citep{berman2000pdb} or AFDB \citep{Varadi2021AFDB}. We uniformly sample the lengths within the supported range (in all our main experiments except long chain generation, the range is set to 50-256) in each training batch and use the mask for loss reweighting. While we are data-free, there is potential to further enhance our method by utilizing these high quality data by adding the adversarial components, we leave that for future study.

\textbf{Time schedule.} During training, we follow the log schedule in \proteina when sampling the timestep~$t$. Choosing $p=2$ and $N_{steps} = 400$, we first sample $s \sim \text{Unif}[0,N_{steps}]$, then define $t$ as:
\begin{equation} \label{eq:tdist}
    t = 1 - 10^{-\frac{s}{N_{steps}}p}
\end{equation}

Generator time steps are equally spaced points on the \emph{discrete index} scale $s \in [s_{init}, N_{steps}]$ of Eq.~\eqref{eq:tdist}, before being converted to continuous time using Eq.~\eqref{eq:tdist}. $s_{init}$ is set to 30 which corresponds to a signal-to-noise-ratio of around 1/2.5, matching the $\sigma_{init}$ value of 2.5 in \citet{zhou2024sid}. Note that $s_{init}$ here is discrete whereas $t_{init}$ in Algorithms~\ref{algo1}--\ref{algo2} lives on the continuous $[0,1]$ scale. The two are related by Eq.~\eqref{eq:tdist}, with $s_{init}=30 \mapsto t_{init}=0.37$.

\textbf{Motif Scaffolding.} The distillation scheme extends naturally for motif scaffolding tasks. Following \proteina \citep{geffner2025proteina}, we randomly sample contiguous blocks from PDB structures \citep{berman2000pdb} as fixed motifs. The rest of the distillation process is identical to unconditional distillation, except that all three networks perform conditional generation on the same motif, and the generator loss is reweighted by scaffold~length. The teacher model for this task is $\mathcal{M}_{\text{motif}}$.

\textbf{Fold Class Conditioning.} For evaluation, we also test our distilled generators' ability to generate structures conditioned on the fold class labels based on the CATH structural hierarchy \citep{Dawson2016CATH}, following the fold class labels defined in \proteina \citep{geffner2025proteina}.

\textbf{Long Chain Generation.} The distillation scheme is identical to the short chain training, except that the range is extended to 50-512. The teacher model for this task is $\mathcal{M}_{\text{long}}$.

\textbf{Parameters for SiD.} The ablation study for $\alpha$ and other parameter settings are discussed in Appendix E. The detailed training procedure is presented in Algo.~\ref{algo1} and sampling procedure in Algo.~\ref{algo2}.

\textbf{Computing resources.} Distillation training is performed on 8 Nvidia A6000-48GB GPUs and takes roughly 4 hours to converge.

\section{Experiments}\label{sec:experiments}
We evaluate SPID by distilling the $\mathcal{M}_\text{FS}^\text{no-tri}$ \proteina model and comparing distilled generators of varying step counts against the teacher across unconditional generation, fold-class conditional generation, and motif scaffolding tasks.
We also validate generalization to an equivariant diffusion architecture by distilling Genie \citep{lin2023genie}.
Throughout, we compare against another distillation method, Diff-Instruct \citep{Luo2024diffinstruct} in all aforementioned generation tasks. Diff-Instruct is a distillation method for image generation. Effectively, its only difference from SiD is the objective: KL divergence instead of Fisher divergence. Therefore, we can adapt it for protein structures in the same way as we do for SiD. Throughout our result sections, "Diff-Instruct" refers to the protein-adapted version, not the original one.
Finally, we perform biological plausibility analyses to verify that the generated protein structures are indeed biologically meaningful.

\subsection{Evaluation}
We adopt the evaluation pipeline from \proteina, which assesses designability, diversity, and novelty through a self-consistency test involving ProteinMPNN \citep{Dauparas2022ProteinMPNN} and ESMFold \citep{Lin2023ESMFold}. A structure is considered designable if the best self-consistency root mean square deviation (scRMSD) is less than 2\r{A}. Designability is the percentage of all generated samples that meet this threshold. Full metric definitions follow \citet{geffner2025proteina}. For each distilled generator, we generate 100 samples for each length in \{50, 100, 150, 200, 250\} and report the designability, diversity, and novelty, as well as the metrics proposed in \proteina (FPSD, fS, and fJSD).
We also record the total generation time across all 500 samples for sampling time analysis.
The lengths in each experiment are specific to the training range. All numbers are single generation runs unless stated otherwise. Designability carries a Wilson 95\% interval; diversity is reported as a point estimate, being a set-level rather than per-structure statistic. See Appendix G for implementation details.

\subsection{Unconditional Generation in One Step}\label{sec:onestep}
We first applied our flow-matching-adjusted score distillation to train a one-step generator. It produced almost no designable structures (0.4\%, Tab.~\ref{tab:main_result}), and a one-step generator distilled with the Diff-Instruct objective under a matched recipe produces none at all (0.0\%). This is consistent with the structural account of Section~\ref{sec:few_step}: at $K=1$ there is no intermediate step at which $\gamma$ could act. We report it as a negative result over a finite set of configurations, not an impossibility.

\subsection{Unconditional Generation in Multiple Steps}\label{sec:uncond_multi}
A critical intermediate finding is that few-step training alone does not recover designability: even with $K > 1$ steps, generators trained via Section~\ref{sec:few_step} result in underwhelming designability under standard sampling (with $\gamma = 1$) as shown in Tab.~\ref{tab:step_gamma_scan}. Only when inference-time noise scaling ($\gamma = 0.45$, Section~\ref{sec:noise_scaling}) is applied does quality improve substantially, indicating that both ingredients of SPID are necessary.

We trained generators with $K \in \{5, 8, 10, 12, 16, 20\}$ steps and reran the pretrained $\mathcal{M}_\text{FS}^\text{no-tri}$ model for a fair comparison.
As shown in Tab.~\ref{tab:main_result} and Fig.~\ref{fig:nstep_discussion} in Appendix H, \emph{designability} improves steadily with step count. Excluding $K=1$ which only produced two designable samples, cluster diversity seems to peak at $K = 8$ (0.76) and is the lowest at $K = 16$ (0.57). 
SPID is statistically indistinguishable from the teacher for $K \geq 10$ --- every difference lies inside the 95\% intervals --- although the point estimates remain 1--2 pp below it until $K = 16$. All SPID generators with matching designability metrics produce slightly less diverse samples than the teacher.
Our distilled generators are also better at generating novel structures distinct from those in the reference databases. Secondary structure content is similar between the 16-step generator and the teacher. We provide a more detailed analysis on the distribution match in Appendix D.

\begin{table*}[t]
    \centering
    \caption{Unconditional backbone generation performance for our distilled generators with different sampling steps compared to the \proteina pretrained model $\mathcal{M}_\text{FS}^\text{no-tri}$ and generators distilled using Diff-Instruct. Optimal performance values are bolded.
    Designability carries a Wilson 95\% interval with $n=500$ for every row. ``Gen.\ time'' is raw wall-clock for all $n$ structures; ``Eff.\ gen.\ time'' divides it by the number of designable samples. Both focus on the time for \emph{backbone generation only}. Diversity, novelty, and effective time are undefined for the SDE rows and the 1-step Diff-Instruct row, which produce no designable structures. Cluster diversity is a ratio over designable samples, so it is not comparable across rows of very different designability. $\gamma$ is set to 0.45 throughout. Behavior at $\gamma = 1$ is reported in Appendix~\ref{app:gamma_frontier}.}
    \resizebox{\textwidth}{!}{%
    \begin{tabular}{l c cc cc cc c cc c cc}
    \vspace{6pt} \\
    \toprule
      Number of & Design- &  \multicolumn{2}{c}{Diversity} & \multicolumn{2}{c}{Novelty}& \multicolumn{2}{c}{FPSD} & fS & \multicolumn{2}{c}{fJSD} & Sec. Struct. \% & Gen. & Eff. gen.\\
      Steps $K$   & ability (\%)$\uparrow$ & Cluster$\uparrow$ & TM$\downarrow$ & PDB$\downarrow$ & AFDB$\downarrow$ & PDB$\downarrow$ & AFDB$\downarrow$ & (C / A / T)$\uparrow$ & PDB$\downarrow$ & AFDB$\downarrow$ & $(\alpha / \beta)$ & time (s) & time (s)\\
    \midrule
     \multicolumn{14}{l}{\emph{SPID}} \\
      1 & 0.4 [0.1, 1.4] &          1.00 (2) &         0.63 &                  0.75 &                  0.76 &  2470.17 &   2403.07 &  2.31 / 2.37 / 4.45 &       2.74 &       2.27 &   75.10 / 0.10 & 11.40 & 5.69\\
      5 & 47.4 [43.1, 51.8] &        0.61 (144) &         0.39 &                  0.75 &                  0.75 &   457.29 &    423.92 & 1.04 / 2.05 / 11.74 &       3.48 &       2.54 &   71.50 / 0.00 & 40.80 & \bf{0.17}\\
      8 & 75.6 [71.6, 79.2] &        \bf{0.76 (286)} &         0.35 &                  0.69 &                  \bf{0.68} &   525.78 &    524.53 & 1.20 / 2.43 / 14.75 &       3.06 &       2.25 &   66.90 / 0.70 & 63.10 & \bf{0.17}\\
     10 & 92.0 [89.3, 94.1] &        0.58 (265) &         0.36 &                  0.71 &                  0.72 &   586.74 &    586.73 & 1.20 / 2.74 / 10.69 &       3.16 &       2.38 &   71.20 / 0.80 & 77.70 & \bf{0.17}\\
     12 & 92.8 [90.2, 94.8] &        0.62 (290) &         0.35 &                  0.71 &                  0.72 &   558.26 &    547.21 & 1.24 / 2.87 / 10.96 &       3.05 &       2.27 &   69.60 / 1.10 & 92.80 & 0.20\\
     16 & 94.4 [92.0, 96.1] &        0.57 (270) &         0.35 &                  0.73 &                  0.74 &   404.04 &    394.20 & 1.86 / 4.27 / 16.18 &       1.94 &       1.54 &   63.20 / 4.60 & 123.20 & 0.26\\
     20 & \bf{95.4} [93.2, 96.9] &        0.58 (277) &         0.35 &                  0.73 &                  0.74 &   515.35 &    510.83 & 1.44 / 3.29 / 13.79 &       2.66 &       2.01 &   68.20 / 2.00 & 152.60 & 0.32\\
    \midrule
     \multicolumn{14}{l}{\emph{Diff-Instruct}} \\
    1 & 0.0 [0.0, 0.8] & --- & --- & --- & --- & 13421.36 & 13548.71 & 1.07 / 1.68 / 2.26 & 4.88 & 5.12 & 26.80 / 9.80 & 11.70 & ---\\
    5 & 4.8 [3.2, 7.0] & 0.62 (15) & 0.39 & \bf{0.67} & \bf{0.68} & 940.74 & 822.64 & 1.93 / 3.55 / 6.40 & 2.78 & 2.49 & 18.30 / 9.50 & 40.60 & 1.69\\
    8 & 19.6 [16.4, 23.3] & 0.68 (67) & 0.37 & 0.73 & 0.75 & 412.89 & 338.12 & 2.32 / 4.70 / 10.92 & 1.71 & 1.57 & 20.90 / 17.70 & 63.60 & 0.65\\
    10 & 34.0 [30.0, 38.3] & 0.62 (105) & 0.36 & 0.75 & 0.78 & 268.03 & 227.38 & 2.50 / 5.76 / 17.31 & 1.00 & 0.98 & 30.20 / 17.30 & 78.00 & 0.46\\
    12 & 43.8 [39.5, 48.2] & 0.62 (136) & 0.34 & 0.75 & 0.79 & 217.94 & 193.40 & 2.41 / \bf{6.11} / 20.75 & 0.85 & 1.01 & 28.70 / 20.90 & 92.80 & 0.42\\
    16 & 64.8 [60.5, 68.9] & 0.67 (218) & \bf{0.33} & 0.75 & 0.78 & 194.72 & 176.90 & 2.54 / 6.01 / 25.51 & \bf{0.65} & 0.69 & 40.20 / 17.70 & 122.50 & 0.38\\
    20 & 78.0 [74.2, 81.4] & 0.67 (261) & \bf{0.33} & 0.76 & 0.79 & 182.27 & 166.43 & 2.46 / 6.01 / 25.58 & \bf{0.65} & 0.66 & 44.40 / 17.80 & 152.10 & 0.39\\
    \midrule
     \multicolumn{14}{l}{\emph{Reduced-step teacher baseline (no distillation)}} \\
    $\mathcal{M}_\text{FS}^\text{no-tri}$, $K{=}10$ (SDE) & 0.0 [0.0, 0.8] & --- & --- & --- & --- & 2035.88 & 1992.35 & 1.56 / 1.76 / 2.03 & 3.26 & 3.24 & 31.5 / 15.8 & 79.0 & ---\\
    $\mathcal{M}_\text{FS}^\text{no-tri}$, $K{=}16$ (SDE) & 0.0 [0.0, 0.8] & --- & --- & --- & --- & 1972.20 & 1899.85 & 2.03 / 2.16 / 2.85 & 3.19 & 3.03 & 16.2 / 27.9 & 123.4 & ---\\
    $\mathcal{M}_\text{FS}^\text{no-tri}$, $K{=}20$ (SDE) & 0.0 [0.0, 0.8] & --- & --- & --- & --- & 2533.66 & 2469.42 & 2.04 / 2.18 / 2.59 & 3.29 & 3.17 & 14.4 / 26.5 & 154.4 & ---\\
    $\mathcal{M}_\text{FS}^\text{no-tri}$, $K{=}40$ (SDE) & 0.0 [0.0, 0.8] & --- & --- & --- & --- & 3565.79 & 3569.90 & 1.93 / 2.45 / 5.31 & 3.61 & 4.00 & 6.4 / 14.4 & 303.0 & ---\\
    $\mathcal{M}_\text{FS}^\text{no-tri}$, $K{=}10$ (ODE) & 1.6 [0.8, 3.1] & 1.00 (8) & 0.35 & 0.71 & 0.70 & 638.01 & 604.03 & 2.49 / 4.01 / 14.71 & 1.76 & 1.71 & 9.60 / 4.00 & 79.00 & 9.88\\
    $\mathcal{M}_\text{FS}^\text{no-tri}$, $K{=}16$ (ODE) & 4.2 [2.8, 6.3] & 0.95 (20) & 0.35 & 0.71 & 0.74 & 330.91 & 336.99 & 2.63 / 4.91 / 30.89 & 1.43 & 1.27 & 21.60 / 2.50 & 123.40 & 5.88\\
    $\mathcal{M}_\text{FS}^\text{no-tri}$, $K{=}20$ (ODE) & 5.2 [3.6, 7.5] & 0.92 (24) & 0.34 & 0.73 & 0.75 & 273.21 & 275.96 & 2.71 / 5.45 / \bf{35.35} & 1.21 & 1.03 & 25.40 / 2.90 & 154.40 & 5.94\\
    $\mathcal{M}_\text{FS}^\text{no-tri}$, $K{=}40$ (ODE) & 7.2 [5.2, 9.8] & 0.89 (32) & 0.35 & 0.74 & 0.75 & \bf{162.07} & \bf{137.44} & \bf{2.81} / 5.46 / 35.13 & 0.89 & \bf{0.55} & 34.80 / 4.30 & 303.00 & 8.42\\
    \midrule
    $\mathcal{M}_\text{FS}^\text{no-tri}$, $K{=}400$ (SDE) & 94.2 [91.8, 95.9] & 0.63 (297) & 0.34 & 0.77 & 0.79 & 316.73 & 290.66 & 1.92 / 4.78 / 20.57 & 1.80 & 1.24 & 64.40 / 4.60 & 3012.00 & 6.40\\
    \bottomrule
    \end{tabular}
    }
    \label{tab:main_result}
\end{table*}

\textbf{Is our recipe necessary?} To separate distillation from the choice of step count, we also run the teacher itself at $K \in \{10,16,20,40\}$ with both its published SDE and ODE solvers. Both solvers result in zero or near-zero designability (Tab.~\ref{tab:main_result}). The SDE sampler uses $g(t) = 1/t$, which diverges as $t \rightarrow 0$, so at large step sizes the noise term tears the chain apart (mean consecutive $C_\alpha$-$C_\alpha$ distance ranges between 21 and 52 \r{A} against the physical 3.8 \r{A}). The ODE sampler produces geometrically valid structures (3.78--3.81 \r{A}) that are nonetheless undesignable. Therefore, it seems reducing the step count alone would not suffice and some form of distillation is required.

\textbf{Against the Diff-Instruct baseline.} At every matched step count SPID is ahead in designability (Tab.~\ref{tab:main_result}). Diff-Instruct is meanwhile ahead on diversity and every distributional metric when $K \geq 10$, which is the same designability--fidelity trade-off that $\gamma$ traces below, although it needs to be noted that the diversity comparison may not be reliable since the designability of Diff-Instruct is much lower.

For the \proteina distributional metrics (FPSD, fS, fJSD), our generators score worse than the teacher at $\gamma=0.45$. However, at $\gamma = 1$ the same generator beats the teacher in terms of the distributional metrics. Sweeping $\gamma$ traces a designability--fidelity frontier (Appendix~\ref{app:gamma_frontier}). Similar to diversity, FPSD, fS, and fJSD can be traded for designability by lowering $\gamma$ until a certain turning point.

\subsection{Backbone Generation Time}\label{sec:sampling_time}
Raw generation time increases linearly with the step count. All three arms, SPID generators, Diff-Instruct generators, and \proteina pretrained generators, have matching raw generation times at a given step count, but differ sharply in generation quality (Tab.~\ref{tab:main_result}). For example, the one-step SPID generator is the fastest in the table in terms of raw generation time, but only 2 of 500 structures are designable.

To account for both generation speed and quality, we also report the \emph{effective} generation time: total time divided by the number of designable structures. This is the last column of Tab.~\ref{tab:main_result} and it is not linear in $K$ --- between $K=5$ and $K=20$ the numerator grows $3.7\times$ while the designable fraction grows $2\times$. 

\textbf{These figures cover backbone generation only.} The designability calculation involves one ProteinMPNN pass and 8 ESMFold predictions per structure and can cost a further 22.68~s per structure. Including it, the cost per designable structure is 24.29~s for SPID against 30.49~s for the teacher: $1.26\times$ end-to-end against $24.5\times$ on the component we distill. We therefore say ``effective backbone-generation time'' throughout and report both. The backbone figure is the one that governs cost as the candidate pool grows or as the designability screen becomes cheaper, which is the regime our introduction motivates.

\subsection{Fold Class Conditional Generation}\label{sec:foldclass}
SPID generalizes to fold-class conditional generation without additional conditional distillation training. 
Using the unconditionally distilled 16-step generator, we provide fold class labels sampled from the empirical joint distribution of length and CATH code \citep{Dawson2016CATH} from AFDB \citep{Varadi2021AFDB} at inference. 
We sample 60 length-CATH-code pairs from the empirical distribution, mask them at the C, A, and T level separately, and ask each of the three models to generate 5 samples for each masked CATH code label, totaling to $n = 900$ per arm. As shown in Tab.~\ref{tab:conditional_result}, the SPID distilled generator is statistically indistinguishable from the teacher ($-3.3$ pp, $p = 0.062$) in terms of designability. The Diff-Instruct arm is 11.0 pp below SPID ($p = 5{\times}10^{-8}$) but ahead on every distributional metric except fS\_A, the same trade-off $\gamma$ traces in Section~\ref{sec:uncond_multi}.

\begin{table*}[t]
    \centering
    \caption{Fold class-conditional generation performance for our distilled generator compared to the \proteina pretrained model $\mathcal{M}_\text{FS}^\text{no-tri}$. Optimal performance values are bolded.
    All three arms share one seeded draw of 60 length-CATH-code pairs; $n = 900$, $\gamma = 0.45$, $K = 16$ throughout. Designability carries a Wilson 95\% interval. The Diff-Instruct arm is distilled under a recipe identical to SPID's, with only the objective changed.}
    \resizebox{\textwidth}{!}{%
    \begin{tabular}{c c cc cc cc c cc c}
    \vspace{6pt} \\
    \toprule
     Model &  Design- &  \multicolumn{2}{c}{Diversity} & \multicolumn{2}{c}{Novelty}& \multicolumn{2}{c}{FPSD} & fS & \multicolumn{2}{c}{fJSD} & Sec. Struct. \% \\
         & ability (\%)$\uparrow$ & Cluster$\uparrow$ & TM$\downarrow$ & PDB$\downarrow$ & AFDB$\downarrow$ & PDB$\downarrow$ & AFDB$\downarrow$ & (C / A / T)$\uparrow$ & PDB$\downarrow$ & AFDB$\downarrow$ & $(\alpha / \beta)$ \\
    \midrule
         16-step SPID & 81.4 [78.8, 83.8] & 0.50 (365) & 0.34 & \bf{0.72} & \bf{0.73} & 239.98 & 268.39 & 2.28 / 6.03 / 21.37 & 1.07 & 1.28 & 50.50 / 11.30 \\
         16-step Diff-Instruct & 70.4 [67.4, 73.3] & \bf{0.62 (391)} & \bf{0.32} & \bf{0.72} & 0.75 & \bf{202.15} & \bf{198.49} & \bf{2.33} / 5.92 / 23.75 & \bf{0.70} & \bf{0.98} & 42.60 / 18.20 \\
         pretrained $\mathcal{M}_\text{FS}^\text{no-tri}$ & \bf{84.7 [82.2, 86.9]} & 0.40 (304) & 0.33 & 0.76 & 0.77 & 237.67 & 262.19 & 2.29 / \bf{6.36} / \bf{23.76} & 0.95 & 1.23 & 49.00 / 13.70 \\
    \bottomrule
    \end{tabular}
    }
    \label{tab:conditional_result}
\end{table*}

\subsection{Long Chain Generation}
One of the strengths of \proteina is its capability of long chain generation, maintaining high designability for proteins of up to 800 residues. To verify that SPID can distill this ability to our generator, we trained a 32-step unconditional generator from \proteina\!\!'s long-chain checkpoint $\mathcal{M}_\text{long}$ and evaluated both the teacher's and the distilled model's long-chain generation performance.
We generate 100 samples per length $L \in \{300, 400, 500\}$ using both the teacher and the distilled generator. We use $\gamma=0.35$ for the teacher model, as suggested by \citet{geffner2025proteina}. For our distilled generator, we set $\gamma = 0.45$ based on a $\gamma$ sweep in Appendix \ref{app:gamma_frontier}. Tab.~\ref{tab:long_chain} shows that our distilled generator has comparable designability and distributional metrics to the teacher for all three lengths, while improving diversity and novelty and significantly reducing the effective generation time. More detailed discussions are in Appendix \ref{app:long_chain}.

\begin{table}[t]
    \centering
    \caption{Long-chain generation: our 32-step generator against \proteina\!\!'s long-chain checkpoint, both measured by us under one evaluation pipeline ($n = 100$ per length). Each model is sampled at its own published operating point --- SPID at $\gamma = 0.45$ with 32 steps, the long-chain checkpoint at $\gamma = 0.35$ with 400 steps, the setting \proteina{} ships it with --- so this is \emph{not} a matched-$\gamma$ comparison; Tab.~\ref{tab:long_chain_gamma} gives the $\gamma$ sensitivity at $L = 500$. Wilson 95\% intervals for designability; effective time is per designable structure.}
    \label{tab:long_chain}
    \resizebox{\textwidth}{!}{%
    \begin{tabular}{lcccccc}
    \vspace{4pt} \\
    \toprule
     & \multicolumn{2}{c}{$L = 300$} & \multicolumn{2}{c}{$L = 400$} & \multicolumn{2}{c}{$L = 500$} \\
    \cmidrule(lr){2-3}\cmidrule(lr){4-5}\cmidrule(lr){6-7}
     & \textbf{SPID} & \proteina $\mathcal{M}_\text{long}$ & \textbf{SPID} & \proteina $\mathcal{M}_\text{long}$ & \textbf{SPID} & \proteina $\mathcal{M}_\text{long}$ \\
    \midrule
    Designability \% $\uparrow$ & \bf{95.0} [88.8, 97.8] & 88.00 [80.2, 93.0] & \bf{93.0} [86.3, 96.6] & 89.00 [81.4, 93.8] & 75.00 [65.7, 82.5] & \bf{79.0} [70.0, 85.8] \\
    Diversity (cluster) $\uparrow$ & \bf{0.66} (63) & 0.45 (40) & \bf{0.73} (68) & 0.49 (44) & \bf{0.76} (57) & 0.70 (55) \\
    Novelty vs.\ PDB $\downarrow$ & \bf{0.73} & 0.82 & \bf{0.69} & 0.81 & \bf{0.66} & 0.81 \\
    PDB-FPSD $\downarrow$ & \bf{602.90} & 669.80 & \bf{694.60} & 741.10 & 953.90 & \bf{948.90} \\
    AFDB-FPSD $\downarrow$ & \bf{543.40} & 613.60 & \bf{621.80} & 668.70 & 890.20 & \bf{877.70} \\
    fS (C / A / T) $\uparrow$ & \bf{2.00} / 4.65 / 8.29 & 1.82 / \bf{4.98} / \bf{9.70} & 1.54 / 3.25 / 8.16 & \bf{1.60 / 4.65 / 9.61} & 1.40 / 3.12 / 4.52 & \bf{1.53 / 3.91 / 5.53} \\
    PDB-fJSD $\downarrow$ & \bf{2.42} & 2.53 & \bf{2.66} & 2.74 & 3.32 & \bf{3.07} \\
    AFDB-fJSD $\downarrow$ & \bf{2.12} & 2.23 & \bf{2.13} & 2.41 & \bf{2.70} & 2.87 \\
    Sec.\ struct. & 59.90 / 5.50 & 59.60 / 4.60 & 67.00 / 3.50 & 61.30 / 4.30 & 67.90 / 2.40 & 61.40 / 4.50 \\
    Effective time (s) $\downarrow$ & \bf{1.49} & 18.64 & \bf{2.49} & 31.00 & \bf{4.42} & 52.33 \\
    \bottomrule
    \end{tabular}
    
    }
\end{table}

\subsection{Motif Scaffolding Benchmark}\label{sec:motif}
We evaluate our 16-step distilled motif scaffolding generator on the RFDiffusion benchmark \citep{watson2023RFDiffusion}, generating 1000 designs per target across all 24 targets with $\gamma=0.4$, the same suggested noise scale for the teacher $\mathcal{M}_\text{motif}$.

In a head-to-head comparison between the two distillation objectives, SPID wins 11 targets, Diff-Instruct 8, with 5 ties. Against our own rerun of the \proteina teacher, SPID is ahead on 17 of 24. Diff-Instruct has \emph{fewer} raw successes than SPID (6140 against 6255) but more unique ones, so its successes are structurally more diverse. Although we cannot conclude which distillation objective is better, it seems that both of them in general produced better motif scaffolding results than our rerun of the teacher, with equal or more unique successes in most of the targets. More detailed analyses are in Appendix \ref{app:motif}.

\subsection{Application to Equivariant Diffusion Architecture}\label{sec:genie}
SPID is not limited to flow-based or non-equivariant models.
To verify generalization to diffusion-based equivariant architectures with IPA and triangle layers, we distill a 10-step generator from the pretrained 1000-step Genie model \citep{lin2023genie}.
The SiD generator loss requires backpropagation through $\xv_g$ into both the fake score network and the teacher; Genie2 \citep{lin2024genie2} includes a one-hot encoding step that blocks this gradient, so we apply SPID to the original Genie instead.
We additionally report scTM-based designability (TM-score threshold 0.5) as defined in \citet{lin2023genie}.

\begin{table*}[t]
    \centering
        \caption{Unconditional generation performance for our distilled generator compared to the Genie pretrained model. $n=100$, 50 samples each for lengths 50 and 100, and $\gamma=0.6$ for both rows. Optimal performance values are bolded.}
        \resizebox{\textwidth}{!}{%
        \begin{tabular}{l cc cc cc cc c cc c}
        \vspace{6pt} \\
        \toprule
         \multirow{2}{*}{Genie Model} & \multicolumn{2}{c}{Designability (\%)} & \multicolumn{2}{c}{Diversity} & \multicolumn{2}{c}{Novelty} & \multicolumn{2}{c}{FPSD} & fS & \multicolumn{2}{c}{fJSD} & Sec. Struct. \% \\
             & scTM $\uparrow$ & scRMSD $\uparrow$ & Cluster$\uparrow$ & TM$\downarrow$ & PDB$\downarrow$ & AFDB$\downarrow$ & PDB$\downarrow$ & AFDB$\downarrow$ & (C / A / T)$\uparrow$ & PDB$\downarrow$ & AFDB$\downarrow$ & $(\alpha / \beta)$ \\
        \midrule
             10-step distilled & \bf{85.0} [76.7, 90.7] & \bf{56.0} [46.2, 65.3] & 0.75 (42) & 0.39 & \bf{0.73} & \bf{0.73} & 476.36 & 411.72 & 1.40 / 2.01 / 9.62 & 2.71 & 2.04 & 60.5 / 2.1 \\
             1000-step pretrained & 84.0 [75.6, 89.9] & 34.0 [25.5, 43.7] & \bf{0.79 (27)} & \bf{0.37} & 0.75 & 0.74 & \bf{328.95} & \bf{282.62} & \bf{2.64 / 4.57 / 15.93} & \bf{1.28} & \bf{0.93} & 41.8 / 10.9 \\
        \bottomrule
        \end{tabular}
        }
        \label{tab:genie_unconditional_result}
\end{table*}

As shown in Tab.~\ref{tab:genie_unconditional_result}, the distilled Genie model achieves a 100$\times$ reduction in sampling steps while improving designability over the 1000-step teacher in terms of scRMSD, although the scTM-based designability only shows insignificant differences. This comes at the cost of slightly worse diversity, FPSD, fS, and fJSD metrics. Similar to the earlier \proteina results, the tradeoff between designability and diversity and the distributional metrics seems to be controlled by $\gamma$ as discussed in Appendix \ref{app:gamma_frontier}. We choose to report the results for $\gamma=0.6$ only to match what the teacher used. The teacher only supports lengths between 50 and 128 (same range for distillation training). Therefore, our evaluation only involved lengths 50 and 100. 

\subsection{Biological Plausibility Analyses}
\begin{figure}
    \centering
    \includegraphics[width=.8\linewidth]{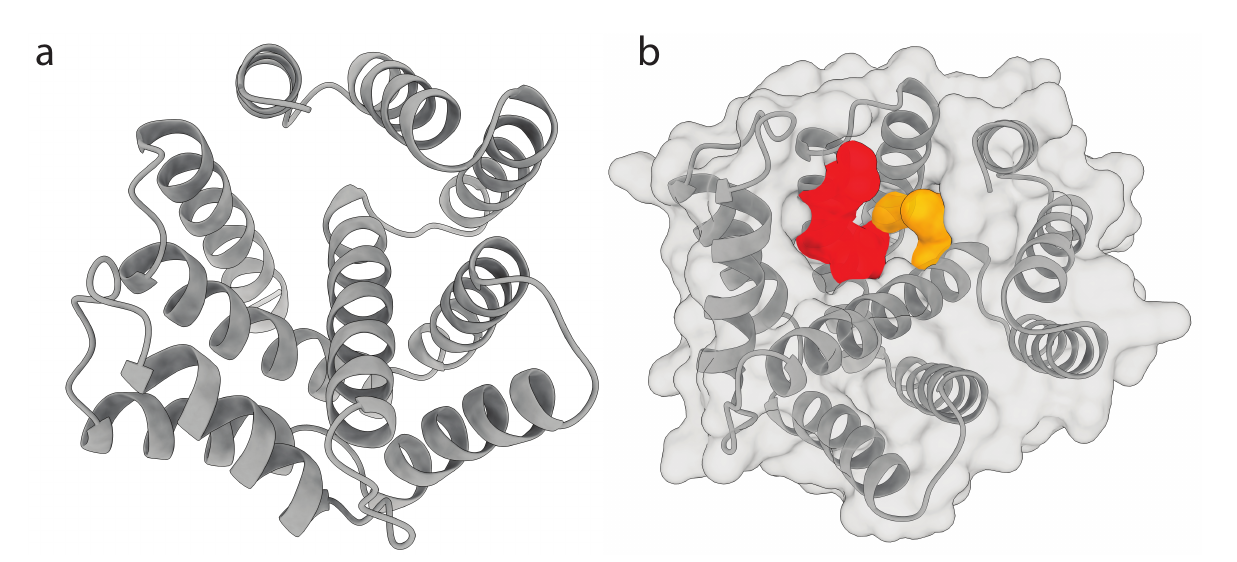}
    \vspace{-3mm}
    \caption{Designed protein backbone and adjacent polar–hydrophobic pocket pair.
(a) Front view of the designed protein backbone.
(b) Cavity view highlighting two adjacent but chemically distinct pockets: Pocket 1 (red, polar) and Pocket 2 (orange, hydrophobic). Two use modes are conceivable: simultaneous binding of a polar and a hydrophobic ligand, or one bifunctional molecule spanning both pockets.}
    \label{fig:case_study}
    \vspace{-3mm}
\end{figure}

To illustrate the structural and functional plausibility of our designed proteins, we conducted a case study on a representative design with high composite confidence (pLDDT = 0.936, pTM = 0.896, RMSD = 0.545), with high novelty score (seqid = 0.14), suggesting that the generated sequence is not a close match to existing entries in the reference database. Fig.~\ref{fig:case_study}a shows the protein backbone, which features a stable helical bundle with well-packed secondary structure. Importantly, automated cavity detection (Fpocket, \citealp{leguilloux2009fpocket}) revealed the presence of two adjacent surface pockets: one predominantly polar and the other largely hydrophobic.

Pocket 1 (red) has a druggability score of $0.758$, a volume of $596$ \AA$^3$, and a polarity score of $7$ (polar-leaning), while Pocket 2 (orange) has a druggability score of $0.796$, a volume of $507$ \AA$^3$, and a polarity score of $3$ (hydrophobic-leaning). These scores suggest both pockets are potential pharmaceutically relevant binding
sites, with sufficient space to accommodate small molecules and complementary chemical environments. On this basis, two binding modes are conceivable: 1) dual-ligand
occupancy, with a polar ligand in Pocket 1 and a hydrophobic ligand in Pocket 2, potentially allowing
cooperative binding; 2) single bifunctional ligand, where a fragment-linked molecule bridges both pockets,
exploiting complementary polar and nonpolar interactions.

We highlight this single case not as evidence that our pipeline reliably generates ligand-binding-competent proteins, but as an illustrative example of the kind of proteins with functionally relevant ligand-binding opportunities that can emerge from our generative pipeline.

\section{Conclusion}\label{sec:conclusion}
We propose SPID, a distillation framework for protein backbone generative models that enables few-step generation with low-temperature sampling, a capability critical for protein design but absent from existing approaches. SPID applies to both diffusion- and flow-based backbone generators and supports data-free training, making it broadly applicable to large pretrained models.

Distilling the flow-based \proteina model without access to training data, SPID generators achieve more than 20$\times$ reduction in effective backbone-generation time while matching the teacher in designability, and remaining comparable in diversity and novelty across unconditional and fold-class conditional generation at lengths 50--256.
A motif scaffolding distillation using real PDB motifs further demonstrates practical utility in targeted design tasks.
SPID also exposes a flexible speed--quality trade-off: though designability increases with step count monotonically, all differences at $K \geq 10$ fall inside their intervals and cluster diversity is not monotone. The degradation in fold-class distributional metrics tracks the inference-time noise scale. A case study on a single structure illustrates cavity geometry. Applying SPID to Genie shows the method is not limited to one model class. Against a Diff-Instruct baseline, the SPID objective's advantage is task-dependent: large on unconditional generation, roughly a third of that under fold-class conditioning, and absent on motif scaffolding.

\textbf{Limitations.} (i) \emph{The step budget scales with problem size}: $K = 16$ for lengths 50--256, $K = 32$ for 300--500, where designability falls to 75.0\% at $L = 500$. Lengths beyond 512 exceed our 48~GB training budget, so we do not inherit \proteina\!\!'s stated capability to 800 residues. (ii) \emph{The speedup is on backbone generation}; end-to-end including the self-consistency pipeline it is $1.26\times$. (iii) \emph{The motif benchmark has an unresolved measurement discrepancy} between our teacher rerun and its published value (Appendix \ref{app:motif}). (iv) SPID is designed for diffusion processes where all noise is Gaussian, as in rectified flow and Gaussian DDPM on coordinates.
Models such as RFDiffusion \citep{watson2023RFDiffusion} and FrameDiff \citep{yim2023framediff} additionally apply Brownian motion on SO(3) to the rotational component of backbone frames; adapting SPID to these models would require extending the score identities to the non-Euclidean geometry of the rotation group. Additionally, models that incorporate discrete or non-differentiable components, such as the one-hot sequence encoding in Genie2 \citep{lin2024genie2}, block the backpropagation required by the SiD generator loss which differentiates with respect to the output of the generator model. Adapting SPID to such architectures would require alternative distillation objectives.

\bibliographystyle{plainnat}
\bibliography{references}

@misc{sabour2025AYF,
      title={Align Your Flow: Scaling Continuous-Time Flow Map Distillation}, 
      author={Amirmojtaba Sabour and Sanja Fidler and Karsten Kreis},
      year={2025},
      eprint={2506.14603},
      archivePrefix={arXiv},
      primaryClass={cs.CV},
      url={https://arxiv.org/abs/2506.14603}, 
}

@misc{boffi2025flowmapmatching,
      title={Flow map matching with stochastic interpolants: A mathematical framework for consistency models}, 
      author={Nicholas M. Boffi and Michael S. Albergo and Eric Vanden-Eijnden},
      year={2025},
      eprint={2406.07507},
      archivePrefix={arXiv},
      primaryClass={cs.LG},
      url={https://arxiv.org/abs/2406.07507}, 
}

@article{shen2025efficient_survey,
    title={Efficient Diffusion Models: A Survey},
    author={Hui Shen and Jingxuan Zhang and Boning Xiong and Rui Hu and Shoufa Chen and Zhongwei Wan and Xin Wang and Yu Zhang and Zixuan Gong and Guangyin Bao and Chaofan Tao and Yongfeng Huang and Ye Yuan and Mi Zhang},
    journal={Transactions on Machine Learning Research},
    issn={2835-8856},
    year={2025},
    url={https://openreview.net/forum?id=wHECkBOwyt},
    note={Survey Certification}
}

@misc{Fan2025survey_distillation,
      title={A Survey on Pre-Trained Diffusion Model Distillations}, 
      author={Xuhui Fan and Zhangkai Wu and Hongyu Wu},
      year={2025},
      eprint={2502.08364},
      archivePrefix={arXiv},
      primaryClass={cs.LG},
      url={https://arxiv.org/abs/2502.08364}, 
}

@misc{zhou2025sidfewstep,
      title={Few-Step Diffusion via Score identity Distillation}, 
      author={Mingyuan Zhou and Yi Gu and Zhendong Wang},
      year={2025},
      eprint={2505.12674},
      archivePrefix={arXiv},
      primaryClass={cs.CV},
      url={https://arxiv.org/abs/2505.12674}, 
}

@inproceedings{geffner2025proteina,
    title={Proteina: Scaling Flow-based Protein Structure Generative Models},
    author={Geffner, Tomas and Didi, Kieran and Zhang, Zuobai and Reidenbach, Danny and Cao, Zhonglin and Yim, Jason and Geiger, Mario and Dallago, Christian and Kucukbenli, Emine and Vahdat, Arash and Kreis, Karsten},
    booktitle={International Conference on Learning Representations (ICLR)},
    year={2025}
}

@inproceedings{zhou2025sida,
    title={Adversarial Score identity Distillation: Rapidly Surpassing the Teacher in One Step},
    author={Mingyuan Zhou and  Huangjie Zheng and Yi Gu and Zhendong Wang and Hai Huang},
    booktitle={International Conference on Learning Representations},
    year={2025},
    url_code={https://github.com/mingyuanzhou/SiD/tree/sida},
    url_arxiv={https://arxiv.org/abs/2410.14919}
}

@misc{kim2024CTM,
      title={Consistency Trajectory Models: Learning Probability Flow ODE Trajectory of Diffusion}, 
      author={Dongjun Kim and Chieh-Hsin Lai and Wei-Hsiang Liao and Naoki Murata and Yuhta Takida and Toshimitsu Uesaka and Yutong He and Yuki Mitsufuji and Stefano Ermon},
      year={2024},
      eprint={2310.02279},
      archivePrefix={arXiv},
      primaryClass={cs.LG},
      url={https://arxiv.org/abs/2310.02279}, 
}

@inproceedings{zhou2024sid,
    title={Score identity Distillation: Exponentially Fast Distillation of Pretrained Diffusion Models for One-Step Generation},
    author={Mingyuan Zhou and Huangjie Zheng and Zhendong Wang and Mingzhang Yin and Hai Huang},
    booktitle={Forty-first International Conference on Machine Learning},
    year={2024},
    url={https://openreview.net/forum?id=QhqQJqe0Wq}
}

@misc{ma2024survey_diffusion,
      title={Efficient Diffusion Models: A Comprehensive Survey from Principles to Practices}, 
      author={Zhiyuan Ma and Yuzhu Zhang and Guoli Jia and Liangliang Zhao and Yichao Ma and Mingjie Ma and Gaofeng Liu and Kaiyan Zhang and Jianjun Li and Bowen Zhou},
      year={2024},
      eprint={2410.11795},
      archivePrefix={arXiv},
      primaryClass={cs.CV},
      url={https://arxiv.org/abs/2410.11795}, 
}

@misc{Bose2024FoldFlow,
	title = {{SE}(3)-{Stochastic} {Flow} {Matching} for {Protein} {Backbone} {Generation}},
	url = {http://arxiv.org/abs/2310.02391},
	doi = {10.48550/arXiv.2310.02391},
	publisher = {arXiv},
	author = {Bose, Avishek Joey and Akhound-Sadegh, Tara and Huguet, Guillaume and Fatras, Kilian and Rector-Brooks, Jarrid and Liu, Cheng-Hao and Nica, Andrei Cristian and Korablyov, Maksym and Bronstein, Michael and Tong, Alexander},
	month = apr,
	year = {2024},
	note = {arXiv:2310.02391 [cs]},
}

@misc{lin2024genie2,
      title={Out of Many, One: Designing and Scaffolding Proteins at the Scale of the Structural Universe with Genie 2}, 
      author={Yeqing Lin and Minji Lee and Zhao Zhang and Mohammed AlQuraishi},
      year={2024},
      eprint={2405.15489},
      archivePrefix={arXiv},
      primaryClass={q-bio.BM},
      url={https://arxiv.org/abs/2405.15489}, 
}

@article{Luo2024diffinstruct,
  title={Diff-instruct: A universal approach for transferring knowledge from pre-trained diffusion models},
  author={Luo, Weijian and Hu, Tianyang and Zhang, Shifeng and Sun, Jiacheng and Li, Zhenguo and Zhang, Zhihua},
  journal={Advances in Neural Information Processing Systems},
  volume={36},
  year={2024}
}

@inproceedings{Wang2024proteus,
    title={Proteus: Exploring Protein Structure Generation for Enhanced Designability and Efficiency},
    author={Chentong Wang and Yannan Qu and Zhangzhi Peng and Yukai Wang and Hongli Zhu and Dachuan Chen and Longxing Cao},
    booktitle={Forty-first International Conference on Machine Learning},
    year={2024},
    url={https://openreview.net/forum?id=IckJCzsGVS}
}

@inproceedings{wang2023vsd,
    title={ProlificDreamer: High-Fidelity and Diverse Text-to-3D Generation with Variational Score Distillation},
    author={Zhengyi Wang and Cheng Lu and Yikai Wang and Fan Bao and Chongxuan Li and Hang Su and Jun Zhu},
    booktitle={Thirty-seventh Conference on Neural Information Processing Systems},
    year={2023},
    url={https://openreview.net/forum?id=ppJuFSOAnM}
}

@inproceedings{lin2023genie,
    author = {Lin, Yeqing and AlQuraishi, Mohammed},
    title = {Generating novel, designable, and diverse protein structures by equivariantly diffusing oriented residue clouds},
    year = {2023},
    publisher = {JMLR.org},
    booktitle = {Proceedings of the 40th International Conference on Machine Learning},
    articleno = {864},
    numpages = {25},
    location = {Honolulu, Hawaii, USA},
    series = {ICML'23}
}

@misc{yim2023frameflow,
	title = {Fast protein backbone generation with {SE}(3) flow matching},
	url = {http://arxiv.org/abs/2310.05297},
	doi = {10.48550/arXiv.2310.05297},
    urldate = {2025-06-03},
	publisher = {arXiv},
	author = {Yim, Jason and Campbell, Andrew and Foong, Andrew Y. K. and Gastegger, Michael and Jiménez-Luna, José and Lewis, Sarah and Satorras, Victor Garcia and Veeling, Bastiaan S. and Barzilay, Regina and Jaakkola, Tommi and Noé, Frank},
	month = oct,
	year = {2023},
	note = {arXiv:2310.05297 [q-bio]}
}

@misc{yim2023framediff,
	title = {{SE}(3) diffusion model with application to protein backbone generation},
	url = {http://arxiv.org/abs/2302.02277},
	doi = {10.48550/arXiv.2302.02277},
	urldate = {2025-06-03},
	publisher = {arXiv},
	author = {Yim, Jason and Trippe, Brian L. and Bortoli, Valentin De and Mathieu, Emile and Doucet, Arnaud and Barzilay, Regina and Jaakkola, Tommi},
	month = may,
	year = {2023},
	note = {arXiv:2302.02277 [cs]}
}

@article{Ingraham2023Chroma,
	title = {Illuminating protein space with a programmable generative model},
	volume = {623},
	copyright = {2023 The Author(s)},
	issn = {1476-4687},
	url = {https://www.nature.com/articles/s41586-023-06728-8},
	doi = {10.1038/s41586-023-06728-8},
	language = {en},
	number = {7989},
	urldate = {2025-09-20},
	journal = {Nature},
	author = {Ingraham, John B. and Baranov, Max and Costello, Zak and Barber, Karl W. and Wang, Wujie and Ismail, Ahmed and Frappier, Vincent and Lord, Dana M. and Ng-Thow-Hing, Christopher and Van Vlack, Erik R. and Tie, Shan and Xue, Vincent and Cowles, Sarah C. and Leung, Alan and Rodrigues, João V. and Morales-Perez, Claudio L. and Ayoub, Alex M. and Green, Robin and Puentes, Katherine and Oplinger, Frank and Panwar, Nishant V. and Obermeyer, Fritz and Root, Adam R. and Beam, Andrew L. and Poelwijk, Frank J. and Grigoryan, Gevorg},
	month = nov,
	year = {2023},
	note = {Publisher: Nature Publishing Group},
	pages = {1070--1078},
}

@article{watson2023RFDiffusion,
	title = {De novo design of protein structure and function with {RFdiffusion}},
	volume = {620},
	copyright = {2023 The Author(s)},
	issn = {1476-4687},
	url = {https://www.nature.com/articles/s41586-023-06415-8},
	doi = {10.1038/s41586-023-06415-8},
	language = {en},
	number = {7976},
	urldate = {2025-06-03},
	journal = {Nature},
	author = {Watson, Joseph L. and Juergens, David and Bennett, Nathaniel R. and Trippe, Brian L. and Yim, Jason and Eisenach, Helen E. and Ahern, Woody and Borst, Andrew J. and Ragotte, Robert J. and Milles, Lukas F. and Wicky, Basile I. M. and Hanikel, Nikita and Pellock, Samuel J. and Courbet, Alexis and Sheffler, William and Wang, Jue and Venkatesh, Preetham and Sappington, Isaac and Torres, Susana Vázquez and Lauko, Anna and De Bortoli, Valentin and Mathieu, Emile and Ovchinnikov, Sergey and Barzilay, Regina and Jaakkola, Tommi S. and DiMaio, Frank and Baek, Minkyung and Baker, David},
	month = aug,
	year = {2023},
	note = {Publisher: Nature Publishing Group},
	pages = {1089--1100}
}

@misc{Trippe2023SMCDiff,
      title={Diffusion probabilistic modeling of protein backbones in 3D for the motif-scaffolding problem}, 
      author={Brian L. Trippe and Jason Yim and Doug Tischer and David Baker and Tamara Broderick and Regina Barzilay and Tommi Jaakkola},
      year={2023},
      eprint={2206.04119},
      archivePrefix={arXiv},
      primaryClass={q-bio.BM},
      url={https://arxiv.org/abs/2206.04119}, 
}

@misc{Song2023ConsistencyModels,
	title = {Consistency {Models}},
	url = {http://arxiv.org/abs/2303.01469},
	doi = {10.48550/arXiv.2303.01469},
	urldate = {2025-06-16},
	publisher = {arXiv},
	author = {Song, Yang and Dhariwal, Prafulla and Chen, Mark and Sutskever, Ilya},
	month = may,
	year = {2023},
	note = {arXiv:2303.01469 [cs]}
}

@article{Lin2023ESMFold, 
    title = {Evolutionary-scale prediction of atomic-level protein structure with a language model}, 
    author = {Zeming Lin and Halil Akin and Roshan Rao and Brian Hie and Zhongkai Zhu and Wenting Lu and Nikita Smetanin and Robert Verkuil and Ori Kabeli and Yaniv Shmueli and Allan dos Santos Costa and Maryam Fazel-Zarandi and Tom Sercu and Salvatore Candido and Alexander Rives }, 
    journal = {Science}, 
    volume = {379}, 
    number = {6637}, 
    pages = {1123-1130}, 
    year = {2023}, 
    doi = {10.1126/science.ade2574}, 
    URL = {https://www.science.org/doi/abs/10.1126/science.ade2574}
}

@inproceedings{Wang2023diffusiongan,
    title={Diffusion-GAN: Training GANs with Diffusion},
    author={Zhendong Wang and Huangjie Zheng and Pengcheng He and Weizhu Chen and Mingyuan Zhou},
    booktitle={The Eleventh International Conference on Learning Representations },
    year={2023},
    url={https://openreview.net/forum?id=HZf7UbpWHuA}
}

@misc{Albergo2023UnifyingFlowDiffusion,
	title = {Stochastic {Interpolants}: {A} {Unifying} {Framework} for {Flows} and {Diffusions}},
	shorttitle = {Stochastic {Interpolants}},
	url = {http://arxiv.org/abs/2303.08797},
	doi = {10.48550/arXiv.2303.08797},
	urldate = {2025-09-23},
	publisher = {arXiv},
	author = {Albergo, Michael S. and Boffi, Nicholas M. and Vanden-Eijnden, Eric},
	month = nov,
	year = {2023},
	note = {arXiv:2303.08797 [cs]},
}

@inproceedings{Liu2023rectifiedFlow,
    title={Flow Straight and Fast: Learning to Generate and Transfer Data with Rectified Flow},
    author={Xingchao Liu and Chengyue Gong and qiang liu},
    booktitle={The Eleventh International Conference on Learning Representations },
    year={2023},
    url={https://openreview.net/forum?id=XVjTT1nw5z}
}

@article{Dauparas2022ProteinMPNN,
    author = {J. Dauparas  and I. Anishchenko  and N. Bennett  and H. Bai  and R. J. Ragotte  and L. F. Milles  and B. I. M. Wicky  and A. Courbet  and R. J. de Haas  and N. Bethel  and P. J. Y. Leung  and T. F. Huddy  and S. Pellock  and D. Tischer  and F. Chan  and B. Koepnick  and H. Nguyen  and A. Kang  and B. Sankaran  and A. K. Bera  and N. P. King  and D. Baker },
    title = {Robust deep learning–based protein sequence design using ProteinMPNN},
    journal = {Science},
    volume = {378},
    number = {6615},
    pages = {49-56},
    year = {2022},
    doi = {10.1126/science.add2187},
    URL = {https://www.science.org/doi/abs/10.1126/science.add2187},
    eprint = {https://www.science.org/doi/pdf/10.1126/science.add2187}
}

@misc{Salimans2022progressivedistillation,
      title={Progressive Distillation for Fast Sampling of Diffusion Models}, 
      author={Tim Salimans and Jonathan Ho},
      year={2022},
      eprint={2202.00512},
      archivePrefix={arXiv},
      primaryClass={cs.LG},
      url={https://arxiv.org/abs/2202.00512}, 
}

@misc{Anand2022Protein_equivariant,
	title = {Protein {Structure} and {Sequence} {Generation} with {Equivariant} {Denoising} {Diffusion} {Probabilistic} {Models}},
	url = {http://arxiv.org/abs/2205.15019},
	doi = {10.48550/arXiv.2205.15019},
	urldate = {2025-09-20},
	publisher = {arXiv},
	author = {Anand, Namrata and Achim, Tudor},
	month = may,
	year = {2022},
	note = {arXiv:2205.15019 [q-bio]},
}

@article{Eguchi2022Ig_VAE,
	title = {Ig-{VAE}: {Generative} modeling of protein structure by direct {3D} coordinate generation},
	volume = {18},
	issn = {1553-7358},
	shorttitle = {Ig-{VAE}},
	url = {https://journals.plos.org/ploscompbiol/article?id=10.1371/journal.pcbi.1010271},
	doi = {10.1371/journal.pcbi.1010271},
	language = {en},
	number = {6},
	urldate = {2025-09-20},
	journal = {PLOS Computational Biology},
	author = {Eguchi, Raphael R. and Choe, Christian A. and Huang, Po-Ssu},
	month = jun,
	year = {2022},
	note = {Publisher: Public Library of Science},
	pages = {e1010271},
}

@article{Varadi2021AFDB,
	title = {{AlphaFold} {Protein} {Structure} {Database}: massively expanding the structural coverage of protein-sequence space with high-accuracy models},
	volume = {50},
	issn = {0305-1048},
	url = {https://doi.org/10.1093/nar/gkab1061},
	doi = {10.1093/nar/gkab1061},
	number = {D1},
	journal = {Nucleic Acids Research},
	author = {Varadi, Mihaly and Anyango, Stephen and Deshpande, Mandar and Nair, Sreenath and Natassia, Cindy and Yordanova, Galabina and Yuan, David and Stroe, Oana and Wood, Gemma and Laydon, Agata and Žídek, Augustin and Green, Tim and Tunyasuvunakool, Kathryn and Petersen, Stig and Jumper, John and Clancy, Ellen and Green, Richard and Vora, Ankur and Lutfi, Mira and Figurnov, Michael and Cowie, Andrew and Hobbs, Nicole and Kohli, Pushmeet and Kleywegt, Gerard and Birney, Ewan and Hassabis, Demis and Velankar, Sameer},
	month = nov,
	year = {2021},
	note = {\_eprint: https://academic.oup.com/nar/article-pdf/50/D1/D439/43502749/gkab1061.pdf},
	pages = {D439--D444},
}

@article{Jumper2021AF2,
	title = {Highly accurate protein structure prediction with {AlphaFold}},
	volume = {596},
	copyright = {2021 The Author(s)},
	issn = {1476-4687},
	url = {https://www.nature.com/articles/s41586-021-03819-2},
	doi = {10.1038/s41586-021-03819-2},
	language = {en},
	number = {7873},
	urldate = {2025-09-20},
	journal = {Nature},
	author = {Jumper, John and Evans, Richard and Pritzel, Alexander and Green, Tim and Figurnov, Michael and Ronneberger, Olaf and Tunyasuvunakool, Kathryn and Bates, Russ and Žídek, Augustin and Potapenko, Anna and Bridgland, Alex and Meyer, Clemens and Kohl, Simon A. A. and Ballard, Andrew J. and Cowie, Andrew and Romera-Paredes, Bernardino and Nikolov, Stanislav and Jain, Rishub and Adler, Jonas and Back, Trevor and Petersen, Stig and Reiman, David and Clancy, Ellen and Zielinski, Michal and Steinegger, Martin and Pacholska, Michalina and Berghammer, Tamas and Bodenstein, Sebastian and Silver, David and Vinyals, Oriol and Senior, Andrew W. and Kavukcuoglu, Koray and Kohli, Pushmeet and Hassabis, Demis},
	month = aug,
	year = {2021},
	note = {Publisher: Nature Publishing Group},
	pages = {583--589},
}

@article{Dawson2016CATH,
	title = {{CATH}: an expanded resource to predict protein function through structure and sequence},
	volume = {45},
	issn = {0305-1048},
	url = {https://doi.org/10.1093/nar/gkw1098},
	doi = {10.1093/nar/gkw1098},
	number = {D1},
	journal = {Nucleic Acids Research},
	author = {Dawson, Natalie L. and Lewis, Tony E. and Das, Sayoni and Lees, Jonathan G. and Lee, David and Ashford, Paul and Orengo, Christine A. and Sillitoe, Ian},
	month = nov,
	year = {2016},
	note = {\_eprint: https://academic.oup.com/nar/article-pdf/45/D1/D289/8846836/gkw1098.pdf},
	pages = {D289--D295},
}

@article{Huang2016protein_design,
	title = {The coming of age of de novo protein design},
	volume = {537},
	copyright = {2016 Springer Nature Limited},
	issn = {1476-4687},
	url = {https://www.nature.com/articles/nature19946},
	doi = {10.1038/nature19946},
	language = {en},
	number = {7620},
	urldate = {2025-06-05},
	journal = {Nature},
	author = {Huang, Po-Ssu and Boyken, Scott E. and Baker, David},
	month = sep,
	year = {2016},
	note = {Publisher: Nature Publishing Group},
	pages = {320--327}
}

@article{leguilloux2009fpocket,
  author     = {Le Guilloux, Vincent and Schmidtke, Peter and Tuff{\'e}ry, Pierre},
  title     = {Fpocket: An open source platform for ligand pocket detection},
  journal   = {BMC Bioinformatics},
  year      = {2009},
  volume    = {10},
  number    = {1},
  pages     = {168},
  doi       = {10.1186/1471-2105-10-168},
  url       = {https://biomedcentral.com}
}

@article{Dougherty2009directed_evolution,
	series = {Protein technologies / {Systems} and synthetic biology},
	title = {Directed evolution: new parts and optimized function},
	volume = {20},
	issn = {0958-1669},
	shorttitle = {Directed evolution},
	url = {https://www.sciencedirect.com/science/article/pii/S0958166909000986},
	doi = {10.1016/j.copbio.2009.08.005},
	number = {4},
	urldate = {2025-06-05},
	journal = {Current Opinion in Biotechnology},
	author = {Dougherty, Michael J and Arnold, Frances H},
	month = aug,
	year = {2009},
	pages = {486--491}
}

@article{Zhang2004TM, 
    title={Scoring function for automated assessment of protein structure template quality}, 
    volume={57}, 
    rights={http://onlinelibrary.wiley.com/termsAndConditions#vor}, ISSN={0887-3585, 1097-0134}, 
    DOI={10.1002/prot.20264}, 
    number={4}, 
    journal={Proteins: Structure, Function, and Bioinformatics}, 
    author={Zhang, Yang and Skolnick, Jeffrey}, 
    year={2004}, 
    month=dec, 
    pages={702–710}, 
    language={en} }

@article{berman2000pdb,
	title = {The {Protein} {Data} {Bank}},
	volume = {28},
	issn = {0305-1048},
	url = {https://doi.org/10.1093/nar/28.1.235},
	doi = {10.1093/nar/28.1.235},
	number = {1},
	journal = {Nucleic Acids Research},
	author = {Berman, Helen M. and Westbrook, John and Feng, Zukang and Gilliland, Gary and Bhat, T. N. and Weissig, Helge and Shindyalov, Ilya N. and Bourne, Philip E.},
	month = jan,
	year = {2000},
	note = {\_eprint: https://academic.oup.com/nar/article-pdf/28/1/235/9895144/280235.pdf},
	pages = {235--242},
}

@article{labesse1997psea,
	title = {P-{SEA}: a new efficient assignment of secondary structure from {C} alpha trace of proteins},
	volume = {13},
	issn = {0266-7061},
	shorttitle = {P-{SEA}},
	doi = {10.1093/bioinformatics/13.3.291},
	language = {eng},
	number = {3},
	journal = {Computer applications in the biosciences: CABIOS},
	author = {Labesse, G. and Colloc'h, N. and Pothier, J. and Mornon, J. P.},
	month = jun,
	year = {1997},
	pmid = {9183534},
	pages = {291--295},
}

\appendix
\section*{Appendix}

\section{Algorithms}
Algorithm \ref{algo1} details the training procedure for SPID.
\begin{algorithm}
\caption{SPID training loop for protein backbone generation} \label{algo1}
\begin{algorithmic}[1]
\STATE \textbf{Input:} Pretrained score network $f_\phi$, generator $G_\theta$, generator score network $f_\psi$, $t_{init}=0.37$, $t_{min}=0.02$, $t_{max}=0.98$, number of generation steps $K=16$, $\alpha=1.0$, optimizer learning rates $\eta_{\psi}=1e^{-4}$ and $\eta_{\theta}=5e^{-5}$.
\STATE \textbf{Initialization: } $\theta \gets \phi, \psi \gets \phi, s_{steps} \gets K\text{ equally spaced points between }30\text{ and }400$
\REPEAT
    \STATE Sample target protein length $N \in \{50, \cdots, 256\}$.
    \IF{The number of generation steps $K = 1$}
        \STATE Sample $\zv \sim \mathcal{N}(0, \mathbf{I})$ and set $\xv_g = G_\theta(t_{\text{init}} \zv)$
    \ELSE
        \STATE Sample $k$ from $\{1, ..., K\}$, then sample $\xv_g$ in $k$ steps recursively using Eq.~\eqref{eq:uniform_step_matching}
    \ENDIF
    \STATE Sample $t$ using Eq.~\eqref{eq:tdist}, and clamp $t$ to be between $t_{min}$ and $t_{max}$.
    \STATE Sample $\epsilonv_t \sim \mathcal{N}(0,\mathbf{I})$. Set $\xv_t = t \text{sg}(\xv_g) + (1-t) \epsilon_t$
    \STATE Update $\psi$ using \eqref{eq:fake_score_loss}:
    \STATE \hspace*{2em}$\mathcal{L}_\psi = \frac{1}{N(1-t)^2}\|f_\psi(x_t, t) - \text{sg}(\xv_g)\|_2^2$, where $N$ is the number of residues.
    \STATE \hspace*{2em}$\psi \gets \psi - \eta_\psi \nabla_\psi \mathcal{L}_\psi$
    \IF{The number of generation steps $K = 1$}
        \STATE Sample $\zv \sim \mathcal{N}(0, \mathbf{I})$ and set $\xv_g = G_\theta(t_{\text{init}} \zv)$
    \ELSE
        \STATE Sample $k$ from $\{1, ..., K\}$, then sample $\xv_g$ in $k$ steps using Eq.~\eqref{eq:uniform_step_matching}
    \ENDIF
    \STATE Sample $t$ using Eq.~\eqref{eq:tdist}, and clamp $t$ to be between $t_{min}$ and $t_{max}$. Compute $\omega(t)$ as defined in Eq.~\eqref{eq:omega_t}.
    \STATE Sample $\epsilonv_t \sim \mathcal{N}(0,\mathbf{I})$. Set $\xv_t = t \xv_g + (1-t) \epsilonv_t$
    \STATE Update $\theta$ using Eq.~\eqref{eq:generator_loss}
    \STATE \hspace*{2em} $\mathcal{L}_\theta = \frac{\omega(t)t^2}{(1-t)^4} [(1-\alpha)\|f_\phi(\xv_t,t)-f_\psi(\xv_t, t)\|_2^2 + (f_\phi(\xv_t,t) - f_\psi(\xv_t,t))^T (f_\psi(\xv_t,t) - \xv_g)]$
    \STATE \hspace*{2em} $\theta \gets \theta - \eta_\theta \nabla_\theta \mathcal{L}_\theta$
    \STATE After every 2 $\theta$-updates, generate 10 samples at random lengths $\in \{50, \cdots, 256\}$ and calculate mean scRMSD.
\UNTIL{The average scRMSD plateaus or the training budget is exhausted}
\STATE \textbf{Output: } $G_\theta$
\STATE \textit{Note: for all our experiments, training is budget-limited to 36k training samples (9 $\theta$-updates) for all teacher models because it is observed that the mean scRMSD in general stops improving after 36k samples. If computational resources allow, one is encouraged to implement an actual plateau check with periodic full 500-sample evaluations.}
\end{algorithmic}
\end{algorithm}

Algorithm \ref{algo2} specifies the inference-time sampling procedure. Note that at inference, Eq.~\eqref{eq:noise_scale_sampling} with noise scale is used instead of the standard Eq.~\eqref{eq:uniform_step_matching} used during training.

\begin{algorithm}
\caption{SPID inference for protein backbone generation} \label{algo2}
\begin{algorithmic}[1]
\STATE \textbf{Input:} Generator $G_\theta$, $t_{init}=0.37$, $\gamma = 0.45$, number of generation steps $K=16$, target protein length $N$.
\STATE \textbf{Initialization: } $s_{steps} \gets K\text{ equally spaced points between }30\text{ and }400$
\IF{The number of generation steps $K = 1$}
        \STATE Sample $\zv \sim \mathcal{N}(0, \mathbf{I})$ and set $\xv_K = G_\theta(t_{\text{init}} \zv)$
\ELSE
    \STATE Sample $\xv_0 \sim \mathcal{N}(0, \mathbf{I})$
    \FOR{$k \in \{0,\cdots,K-1\}$}
        \STATE Sample $\xv_{k+1}$ from $\xv_k$ using Eq.~\eqref{eq:noise_scale_sampling} with noise scale $\gamma$ and time path $s_{steps}$, taking $s_{k+1}$ as the $(k{+}1)$-th entry of $s_{steps}$
    \ENDFOR
\ENDIF

\STATE \textbf{Output: } $\xv_K$
\end{algorithmic}
\end{algorithm}

\section{Effect of Sampling Noise Scale}\label{app:gamma_frontier}
\begin{figure}
\vskip 0.2in
\centering
\includegraphics[width=0.7\linewidth]{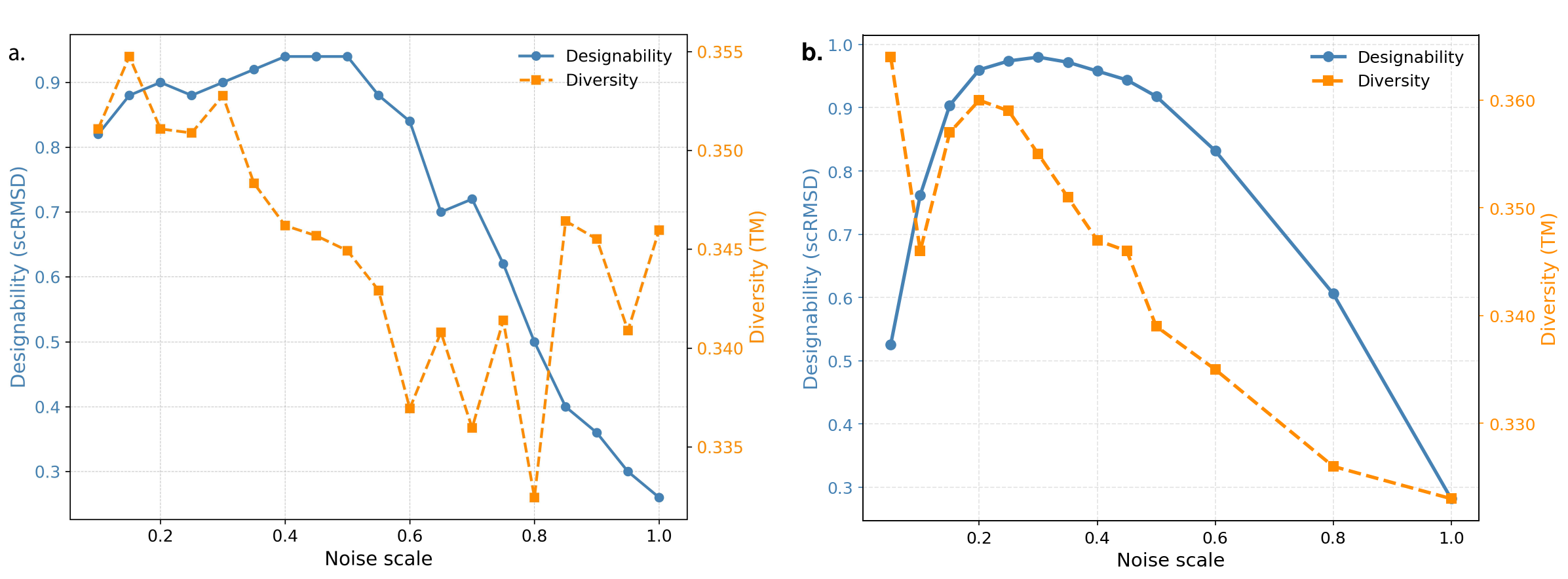}
\caption{An illustration of the effects of the noise scale on the designability (scRMSD) and diversity (TM) of our a. 10-step generator and b. 16-step generator. For designability, it is the fraction of samples in the batch with scRMSD $<$ 2\r{A}. The higher the designability the better. For the diversity metric, it is measuring the average similarity (TM-score) within the batch. Lower values indicate more diversity. Setting the noise scale to 1 is the standard way of sampling in diffusion and flow matching models for images, which results in significantly lower designability for protein structures. For the 10-step generator the designability is the highest for noise scales around 0.45 while the best diversity is reached at 0.8. For the 16-step generator, the highest designability is achieved at 0.3 while the best diversity is reached at 1. $\gamma=0.45$ has the best diversity score among all values that achieve a designability close to the teacher (in point estimate).}
\label{fig:noise_scale_discussion}
\end{figure}

As discussed by prior works, the noise scale during sampling is crucial for protein structure generation tasks. We empirically verify the impact of noise scaling and determine our optimal noise scale.
The sweep in Fig.~\ref{fig:noise_scale_discussion} was run on the 10-step and 16-step generators with 500-sample evaluation sets and lengths in $\{50, 100, 150, 200, 250\}$.

As shown in Fig.~\ref{fig:noise_scale_discussion}, different step counts result in different designability peaks. For the 10-step generator, the designability seems to peak at around $\gamma=0.45$ whereas for the 16-step generator, the designability peaks around $\gamma=0.3$. In both cases, $\gamma=0.45$ achieves the best balance between designability and diversity, which is why we choose 0.45 for our main results.
The $\gamma$ value that we pick coincides with the value picked for the $\mathcal{M}_\text{FS}^\text{no-tri}$ model in \proteina, but $\gamma$ can be tuned to prioritize different metrics.

\begin{table}[t]
    \centering
    \caption{The designability--fidelity frontier for the \textbf{16-step} generator, over $\gamma \in [0.05, 1.00]$. $n = 500$ (100 samples each for lengths $\{50, 100, 150, 200, 250\}$) for every row. Designability carries a Wilson 95\% interval.}
    \label{tab:gamma_sweep_proteina}
    \resizebox{\textwidth}{!}{%
    \begin{tabular}{l c cc cc cc c cc c}
    \vspace{4pt} \\
    \toprule
    \multirow{2}{*}{$\gamma$} & Design- & \multicolumn{2}{c}{Diversity} & \multicolumn{2}{c}{Novelty} & \multicolumn{2}{c}{FPSD} & fS & \multicolumn{2}{c}{fJSD} & Sec. Struct. \% \\
     & ability (\%)$\uparrow$ & Cluster$\uparrow$ & TM$\downarrow$ & PDB$\downarrow$ & AFDB$\downarrow$ & PDB$\downarrow$ & AFDB$\downarrow$ & (C / A / T)$\uparrow$ & PDB$\downarrow$ & AFDB$\downarrow$ & $(\alpha / \beta)$ \\
    \midrule
    \multicolumn{12}{l}{\emph{16-step distilled \proteina}} \\
    0.05 & 52.6 [48.2, 56.9] & 0.32 (85) & 0.364 & \bf{0.684} & \bf{0.684} & 716.72 & 669.76 & 2.24 / 4.77 / 6.41 & 3.292 & 2.907 & 39.8 / 5.0 \\
    0.10 & 76.2 [72.3, 79.7] & 0.39 (148) & 0.346 & 0.713 & 0.719 & 617.55 & 580.18 & 2.20 / 5.07 / 8.20 & 2.981 & 2.517 & 49.3 / 4.3 \\
    0.15 & 90.4 [87.5, 92.7] & 0.44 (198) & 0.357 & 0.732 & 0.733 & 622.99 & 598.20 & 1.85 / 4.50 / 8.57 & 2.895 & 2.349 & 61.0 / 2.4 \\
    0.20 & 96.0 [93.9, 97.4] & 0.45 (217) & 0.360 & 0.731 & 0.733 & 650.50 & 632.67 & 1.62 / 3.87 / 8.59 & 2.875 & 2.286 & 66.5 / 1.6 \\
    0.25 & 97.4 [95.6, 98.5] & 0.47 (229) & 0.359 & 0.730 & 0.729 & 631.09 & 614.74 & 1.52 / 3.72 / 9.40 & 2.838 & 2.215 & 68.4 / 1.8 \\
    0.30 & \bf{98.0} [96.4, 98.9] & 0.51 (250) & 0.355 & 0.732 & 0.733 & 571.30 & 559.76 & 1.52 / 3.57 / 10.46 & 2.761 & 2.149 & 68.3 / 2.0 \\
    0.35 & 97.2 [95.4, 98.3] & 0.52 (255) & 0.351 & 0.732 & 0.734 & 514.10 & 503.91 & 1.57 / 3.66 / 12.44 & 2.535 & 1.978 & 67.4 / 2.4 \\
    0.40 & 95.8 [93.7, 97.2] & 0.58 (278) & 0.347 & 0.724 & 0.730 & 464.03 & 457.83 & 1.63 / 3.85 / 14.24 & 2.296 & 1.769 & 66.1 / 3.2 \\
    0.45 & 94.4 [92.0, 96.1] & 0.57 (270) & 0.346 & 0.733 & 0.741 & 404.04 & 394.20 & 1.86 / 4.27 / 16.18 & 1.94 & 1.54 & 63.2 / 4.6 \\
    0.50 & 91.8 [89.1, 93.9] & 0.64 (294) & 0.339 & 0.727 & 0.727 & 360.12 & 357.65 & 1.77 / 4.13 / 17.71 & 1.865 & 1.405 & 62.6 / 4.9 \\
    0.60 & 83.2 [79.7, 86.2] & 0.70 (291) & 0.335 & 0.730 & 0.740 & 274.47 & 274.41 & 1.97 / 4.55 / 22.70 & 1.407 & 1.046 & 58.5 / 6.5 \\
    0.80 & 60.6 [56.3, 64.8] & 0.74 (225) & 0.326 & 0.730 & 0.743 & 162.87 & 160.24 & 2.26 / 5.11 / 29.37 & 0.826 & 0.647 & 48.8 / 9.2 \\
    1.00 & 28.2 [24.4, 32.3] & \bf{0.81} (114) & \bf{0.323} & 0.741 & 0.750 & \bf{114.01} & \bf{96.06} & \bf{2.55 / 6.13 / 35.96} & \bf{0.467} & \bf{0.461} & 36.7 / 10.2 \\
    \midrule
    \emph{Teacher, 400 steps, $\gamma = 0.45$} & \emph{94.2 [91.8, 95.9]} & \emph{0.63 (297)} & \emph{0.338} & \emph{0.772} & \emph{0.786} & \emph{316.73} & \emph{290.66} & \emph{1.92 / 4.78 / 20.57} & \emph{1.80} & \emph{1.24} & \emph{64.4 / 4.6} \\
    \bottomrule
    \end{tabular}
    }
\end{table}

\begin{table}[t]
    \centering
    \caption{Noise-scale sensitivity of the 32-step long-chain generator at $L = 500$. Samples size $n = 100$ throughout; designability carries a Wilson 95\% interval. We report $\gamma = 0.45$ because it is the value used for short chains, not because it is the optimum. The best cluster diversity is not bolded because it is unfair to compare when the designability is drastically different.}
    \label{tab:long_chain_gamma}
    \resizebox{\textwidth}{!}{%
    \begin{tabular}{l c cc cc cc c cc c}
    \vspace{4pt} \\
    \toprule
    \multirow{2}{*}{$\gamma$} & Design- & \multicolumn{2}{c}{Diversity} & \multicolumn{2}{c}{Novelty} & \multicolumn{2}{c}{FPSD} & fS & \multicolumn{2}{c}{fJSD} & Sec. Struct. \% \\
     & ability (\%)$\uparrow$ & Cluster$\uparrow$ & TM$\downarrow$ & PDB$\downarrow$ & AFDB$\downarrow$ & PDB$\downarrow$ & AFDB$\downarrow$ & (C / A / T)$\uparrow$ & PDB$\downarrow$ & AFDB$\downarrow$ & $(\alpha / \beta)$ \\
    \midrule
    0.20 & 14.0 [8.4, 22.5] & 0.79 (11) & 0.414 & 0.657 & 0.599 & 2335.0 & 2261.3 & 1.91 / 2.18 / 2.20 & 4.746 & 4.565 & 33.3 / 1.3 \\
    0.25 & 21.0 [14.0, 30.3] & 0.81 (17) & 0.390 & 0.695 & 0.652 & 2251.4 & 2176.8 & 1.78 / 2.13 / 2.03 & 4.759 & 4.573 & 36.9 / 1.5 \\
    0.30 & 46.0 [36.6, 55.7] & 0.65 (30) & 0.352 & 0.696 & 0.654 & 1801.7 & 1734.2 & 2.03 / 2.82 / 2.82 & 4.158 & 3.855 & 47.4 / 1.4 \\
    0.35 & 68.0 [58.3, 76.3] & 0.65 (44) & 0.337 & 0.677 & 0.634 & 1446.4 & 1384.6 & 1.82 / 3.24 / 3.53 & 3.856 & 3.365 & 58.5 / 1.7 \\
    0.40 & 76.0 [66.8, 83.3] & 0.76 (58) & 0.329 & 0.677 & 0.652 & 1100.2 & 1041.7 & 1.57 / 3.30 / 4.43 & 3.534 & 2.958 & 63.8 / 2.3 \\
    0.45 & 75.0 [65.7, 82.5] & 0.76 (57) & 0.328 & 0.659 & 0.631 & 953.9 & 890.2 & 1.40 / 3.12 / 4.52 & 3.323 & 2.701 & 67.9 / 2.4 \\
    0.50 & \bf{79.0} [70.0, 85.8] & 0.80 (63) & 0.319 & 0.663 & 0.634 & 816.6 & 757.9 & 1.40 / 3.28 / 6.16 & 2.826 & 2.252 & 66.7 / 3.7 \\
    0.60 & 57.0 [47.2, 66.3] & 0.88 (50) & 0.307 & 0.645 & 0.626 & 739.0 & 692.3 & 1.47 / 3.24 / 7.11 & 2.495 & 1.958 & 63.1 / 5.2 \\
    0.70 & 48.0 [38.5, 57.7] & 0.98 (47) & 0.302 & \bf{0.613} & \bf{0.580} & 663.4 & 620.0 & 1.57 / 3.19 / 7.34 & 2.101 & 1.612 & 59.3 / 5.4 \\
    0.80 & 22.0 [15.0, 31.1] & 1.00 (22) & \bf{0.301} & 0.624 & 0.611 & 576.1 & 537.7 & 1.81 / \bf{4.31} / 11.27 & 1.762 & 1.368 & 54.2 / 6.0 \\
    0.90 & 4.0 [1.6, 9.8] & 1.00 (4) & 0.302 & 0.608 & 0.521 & 529.1 & 501.4 & 1.77 / 3.51 / 9.83 & 1.550 & 1.251 & 49.1 / 5.7 \\
    1.00 & 2.0 [0.6, 7.0] & 1.00 (2) & 0.323 & 0.612 & 0.556 & \bf{486.3} & \bf{457.8} & \bf{1.88} / 3.87 / \bf{12.34} & \bf{1.290} & \bf{1.063} & 42.3 / 5.2 \\
    \bottomrule
    \end{tabular}
    }
\end{table}

\begin{table}[t]
    \centering
    \caption{Effect of the noise scale on the 10-step distilled Genie generator. $n = 100$ at lengths 50 and 100 for every row. Bold marks the best value among the distilled rows.}
    \label{tab:gamma_sweep_genie}
    \resizebox{\textwidth}{!}{%
    \begin{tabular}{l cc cc cc cc c cc c}
    \vspace{4pt} \\
    \toprule
    \multirow{2}{*}{$\gamma$} & \multicolumn{2}{c}{Designability (\%)} & \multicolumn{2}{c}{Diversity} & \multicolumn{2}{c}{Novelty} & \multicolumn{2}{c}{FPSD} & fS & \multicolumn{2}{c}{fJSD} & Sec. Struct. \% \\
     & scTM $\uparrow$ & scRMSD $\uparrow$ & Cluster$\uparrow$ & TM$\downarrow$ & PDB$\downarrow$ & AFDB$\downarrow$ & PDB$\downarrow$ & AFDB$\downarrow$ & (C / A / T)$\uparrow$ & PDB$\downarrow$ & AFDB$\downarrow$ & $(\alpha / \beta)$ \\
    \midrule
    \multicolumn{13}{l}{\emph{10-step distilled Genie}} \\
    0.30 & \bf{90.0} [82.6, 94.5] & \bf{69.0} [59.4, 77.2] & 0.70 (48) & 0.393 & 0.712 & 0.711 & 568.86 & 496.68 & 1.15 / 1.42 / 8.36 & 3.465 & 2.613 & 63.3 / 0.4 \\
    0.45 & 89.0 [81.4, 93.7] & 58.0 [48.2, 67.2] & 0.72 (42) & 0.391 & 0.720 & 0.712 & 530.37 & 458.08 & 1.23 / 1.69 / 9.35 & 3.170 & 2.324 & 62.4 / 0.6 \\
    0.60 & 85.0 [76.7, 90.7] & 56.0 [46.2, 65.3] & 0.75 (42) & 0.386 & 0.726 & 0.733 & 476.36 & 411.72 & 1.40 / 2.01 / 9.62 & 2.711 & 2.040 & 60.5 / 2.1 \\
    0.80 & 84.0 [75.6, 89.9] & 46.0 [36.6, 55.7] & 0.83 (38) & \bf{0.367} & \bf{0.676} & \bf{0.703} & 450.47 & 394.45 & 1.68 / 2.29 / 11.86 & 2.248 & 1.698 & 54.1 / 3.2 \\
    1.00 & 76.0 [66.8, 83.3] & 39.0 [30.0, 48.8] & \bf{0.87} (34) & 0.377 & 0.728 & 0.717 & \bf{414.15} & \bf{365.80} & \bf{2.34 / 3.44 / 15.13} & \bf{1.614} & \bf{1.283} & 46.8 / 5.0 \\
    \midrule
    \emph{Teacher, 1000 steps} & \emph{84.0 [75.6, 89.9]} & \emph{34.0 [25.5, 43.7]} & \emph{0.79 (27)} & \emph{0.367} & \emph{0.751} & \emph{0.739} & \emph{328.95} & \emph{282.62} & \emph{2.64 / 4.57 / 15.93} & \emph{1.277} & \emph{0.928} & \emph{41.8 / 10.9} \\
    \bottomrule
    \end{tabular}
    }
\end{table}

Tab.~\ref{tab:gamma_sweep_proteina}, Tab.~\ref{tab:long_chain_gamma}, and Tab.~\ref{tab:gamma_sweep_genie} present the $\gamma$ sweep on both \proteina (short and long chain) and Genie as teachers. All three tables suggest that lower $\gamma$ values in general lead to better designability but worse diversity and the distributional metrics, although very low $\gamma$ values seem to degrade designability without recovering diversity and the distributional metrics according to the \proteina sweeps.

\section{Motif Scaffolding Results} \label{app:motif}
Motif scaffolding results are generated in the same way as described in the \citet{geffner2025proteina} paper. We generated 1000 designs per target in the RFDiffusion benchmark set \citep{watson2023RFDiffusion}. For each backbone design, 8 sequences are generated by ProteinMPNN and fed to ESMFold. The scRMSD is then computed using the predicted backbones. The motifRMSD is the RMSD between the predicted backbone and the ground truth motif. Successful designs are filtered out using the criteria: scRMSD $\leq$ 2\r{A}, motifRMSD $\leq$ 1\r{A}, pLDDT $\geq$ 70, and pAE $\leq$ 5. The successful designs are then clustered via hierarchical clustering with single linkage and a TM-score
threshold of 0.6 to reach the final number of unique successes, which are reported in Tab.~\ref{tab:motif_scaffolding}.

\begin{table}[t]
\centering
\caption{Motif scaffolding RFDiffusion benchmark: unique successes out of 1{,}000 designs per target. $\gamma$ is set to 0.4 for the three arms we measured. The four leftmost columns are imported from \citet{geffner2025proteina} --- \proteina, Genie2~\citep{lin2024genie2}, RFDiffusion~\citep{watson2023RFDiffusion} and FrameFlow~\citep{yim2023frameflow}. The next three were measured by us with a single pipeline and clustering implementation under the same criterion and are mutually comparable. Bold marks the best of those three per target. \textbf{Bold does not mark the best method overall}: our rerun of the \proteina teacher (488) and its imported value (2094) differ significantly under what should be identical metrics, so the imported and measured blocks cannot safely be ranked against each other. Rows are ordered by SPID's count; we report no aggregate ranking (Section~\ref{sec:motif}).}
\label{tab:motif_scaffolding}
\setlength{\tabcolsep}{4pt}
\resizebox{\textwidth}{!}{%
\begin{tabular}{lrrrr|rrr}
\toprule
 & \multicolumn{4}{c|}{\emph{Imported}} & \multicolumn{3}{c}{\emph{Measured by us}} \\
\textbf{Target} & \textbf{\proteina} & \textbf{Genie2} & \textbf{RFDiff} & \textbf{FrameFlow} & \textbf{SPID} & \textbf{Diff-Instruct} & \textbf{\proteina}\\
\midrule
1QJG          &    3 &    5 &    1 &   18 & 386 & \textbf{407} & 28\\
6E6R\_long    &  713 &  415 &  381 &  110 & \textbf{376} & 369 & 126 \\
6EXZ\_long    &  290 &  326 &  167 &  403 & 277 & \textbf{321} & 45 \\
5TRV\_long    &  179 &   97 &   23 &   77 & 214 & \textbf{235} & 26 \\
6E6R\_medium  &  417 &  272 &  151 &   99 & 120 & \textbf{155} & 26 \\
7MRX\_128     &   51 &   27 &   66 &   35 & \textbf{109} & 96 & 104 \\
6EXZ\_medium  &   43 &   54 &   25 &  110 & \textbf{70} & 65 & 5 \\
1YCR          &  249 &  134 &    7 &  149 & \textbf{54} & 36 & 21 \\
5TRV\_medium  &   22 &   23 &   10 &   21 & 31 & \textbf{41} & 10 \\
5YUI          &    5 &    3 &    1 &    1 & \textbf{27} & \textbf{27} & 13 \\
6EXZ\_short   &    3 &    2 &    1 &    3 & 15 & \textbf{21} & 7 \\
7MRX\_85      &   31 &   23 &   13 &   22 & \textbf{13} & 9 & 12 \\
5TPN          &    4 &    8 &    5 &    6 & 11 & \textbf{13} & 5 \\
6E6R\_short   &   56 &   26 &   23 &   25 & 11 & 14 & \textbf{23} \\
4JHW          &    0 &    0 &    0 &    0 & \textbf{10} & 9 & 9 \\
3IXT          &    8 &   14 &    3 &    8 & \textbf{9} & 5 & 7 \\
4ZYP          &   11 &    3 &    6 &    4 & \textbf{7} & 3 & 5 \\
5TRV\_short   &    1 &    3 &    1 &    1 & \textbf{5} & 4 & 1 \\
5IUS          &    1 &    1 &    1 &    0 & 2 & 1 & \textbf{4} \\
1PRW          &    1 &    1 &    1 &    1 & \textbf{1} & \textbf{1} & \textbf{1} \\
2KL8          &    1 &    1 &    1 &    1 & \textbf{1} & 0 & \textbf{1} \\
7MRX\_60      &    2 &    5 &    1 &    1 & 1 & 1 & \textbf{7} \\
1BCF          &    1 &    1 &    1 &    1 & \textbf{1} & \textbf{1} & \textbf{1} \\
5WN9          &    2 &    1 &    0 &    3 & \textbf{1} & \textbf{1} & \textbf{1} \\
\midrule
\textbf{Total} & 2094 & 1445 & 889 & 1099 & 1752 & \textbf{1835} & 488 \\
\bottomrule
\end{tabular}
}
\end{table}

It is unclear to us why our \proteina teacher rerun is so different from the imported numbers. However, with the same evaluation pipeline, both distilled generators produced significantly more unique successes than the \proteina teacher in targets including 1QJG, 6E6R\_long, 6EXZ\_long, 5TRV\_long, 6E6R\_medium, and 6EXZ\_medium. To further investigate, Tab.~\ref{tab:motif_raw_vs_unique} gives the raw success count $S$ alongside the unique count $U$ for the three arms we measured, together with $r = U/S$, the fraction of raw successes that survive clustering. The teacher posts \emph{more} raw successes than SPID on 22 of 24 targets and ends with fewer unique ones, because $r$ is an order of magnitude smaller for it (pooled ratios 0.046 against 0.280). The \proteina teacher's raw successes are concentrated on repeated solutions. We hypothesize that this is a sign of the distilled generators achieving higher diversity. A closer inspection of the successful designs would be needed to settle this.

\begin{table}[t]
\centering
\caption{Raw versus unique successes on the motif-scaffolding benchmark, 1{,}000 designs per target, $\gamma=0.4$, for the three arms measured with a single pipeline. $S$ is the number of successful designs; $U$ is the number of unique successes after hierarchical clustering with single linkage at TM 0.6, as in Tab.~\ref{tab:motif_scaffolding}; $r = U/S$. Low $r$ means an arm keeps rediscovering the same solutions. Note the targets where $r$ is near zero for every arm (1BCF, 3IXT, 5TRV\_short), on which no arm produces more than a handful of distinct scaffolds and the comparison carries little information; and 7MRX\_85 and 6E6R\_short, where SPID's $r$ is high but its raw count is small, so a large per-design advantage yields almost no margin. Rows in the order of Tab.~\ref{tab:motif_scaffolding}.}
\label{tab:motif_raw_vs_unique}
\setlength{\tabcolsep}{4pt}
\resizebox{\textwidth}{!}{%
\begin{tabular}{lrrr|rrr|rrr}
\toprule
 & \multicolumn{3}{c|}{\textbf{SPID}} & \multicolumn{3}{c|}{\textbf{Diff-Instruct}} & \multicolumn{3}{c}{\textbf{\proteina teacher}} \\
\textbf{Target} & $S$ & $U$ & $r$ & $S$ & $U$ & $r$ & $S$ & $U$ & $r$ \\
\midrule
1QJG          & 856 & 386 & 0.451 & 892 & 407 & 0.456 & 974 &  28 & 0.029 \\
6E6R\_long    & 463 & 376 & 0.812 & 431 & 369 & 0.856 & 893 & 126 & 0.141 \\
6EXZ\_long    & 583 & 277 & 0.475 & 589 & 321 & 0.545 & 938 &  45 & 0.048 \\
5TRV\_long    & 544 & 214 & 0.393 & 498 & 235 & 0.472 & 968 &  26 & 0.027 \\
6E6R\_medium  & 208 & 120 & 0.577 & 236 & 155 & 0.657 & 535 &  26 & 0.049 \\
7MRX\_128     & 120 & 109 & 0.908 & 106 &  96 & 0.906 & 753 & 104 & 0.138 \\
6EXZ\_medium  & 486 &  70 & 0.144 & 463 &  65 & 0.140 & 747 &   5 & 0.007 \\
1YCR          & 265 &  54 & 0.204 & 254 &  36 & 0.142 & 580 &  21 & 0.036 \\
5TRV\_medium  & 467 &  31 & 0.066 & 462 &  41 & 0.089 & 798 &  10 & 0.013 \\
5YUI          & 190 &  27 & 0.142 & 186 &  27 & 0.145 & 392 &  13 & 0.033 \\
6EXZ\_short   & 117 &  15 & 0.128 & 103 &  21 & 0.204 & 278 &   7 & 0.025 \\
7MRX\_85      &  19 &  13 & 0.684 &  11 &   9 & 0.818 & 277 &  12 & 0.043 \\
5TPN          &  24 &  11 & 0.458 &  26 &  13 & 0.500 &  34 &   5 & 0.147 \\
6E6R\_short   &  15 &  11 & 0.733 &  20 &  14 & 0.700 & 137 &  23 & 0.168 \\
4JHW          &  40 &  10 & 0.250 &  23 &   9 & 0.391 & 104 &   9 & 0.087 \\
3IXT          & 376 &   9 & 0.024 & 402 &   5 & 0.012 & 634 &   7 & 0.011 \\
4ZYP          &  64 &   7 & 0.109 &  45 &   3 & 0.067 &  99 &   5 & 0.051 \\
5TRV\_short   & 238 &   5 & 0.021 & 226 &   4 & 0.018 & 443 &   1 & 0.002 \\
5IUS          & 203 &   2 & 0.010 & 198 &   1 & 0.005 & 291 &   4 & 0.014 \\
1PRW          &   5 &   1 & 0.200 &   1 &   1 & 1.000 &  41 &   1 & 0.024 \\
2KL8          &   1 &   1 & 1.000 &   0 &   0 & --    &   1 &   1 & 1.000 \\
7MRX\_60      &   1 &   1 & 1.000 &   1 &   1 & 1.000 &  10 &   7 & 0.700 \\
1BCF          & 879 &   1 & 0.001 & 877 &   1 & 0.001 & 667 &   1 & 0.001 \\
5WN9          &  91 &   1 & 0.011 &  90 &   1 & 0.011 & 112 &   1 & 0.009 \\
\midrule
\textbf{Total} & \textbf{6255} & \textbf{1752} & \textbf{0.280} & \textbf{6140} & \textbf{1835} & \textbf{0.299} & \textbf{10706} & \textbf{488} & \textbf{0.046} \\
\bottomrule
\end{tabular}
}
\end{table}

\section{Secondary Structure Distributions}
\begin{figure}
    \vskip 0.2in
    \centering
    \includegraphics[width=\linewidth]{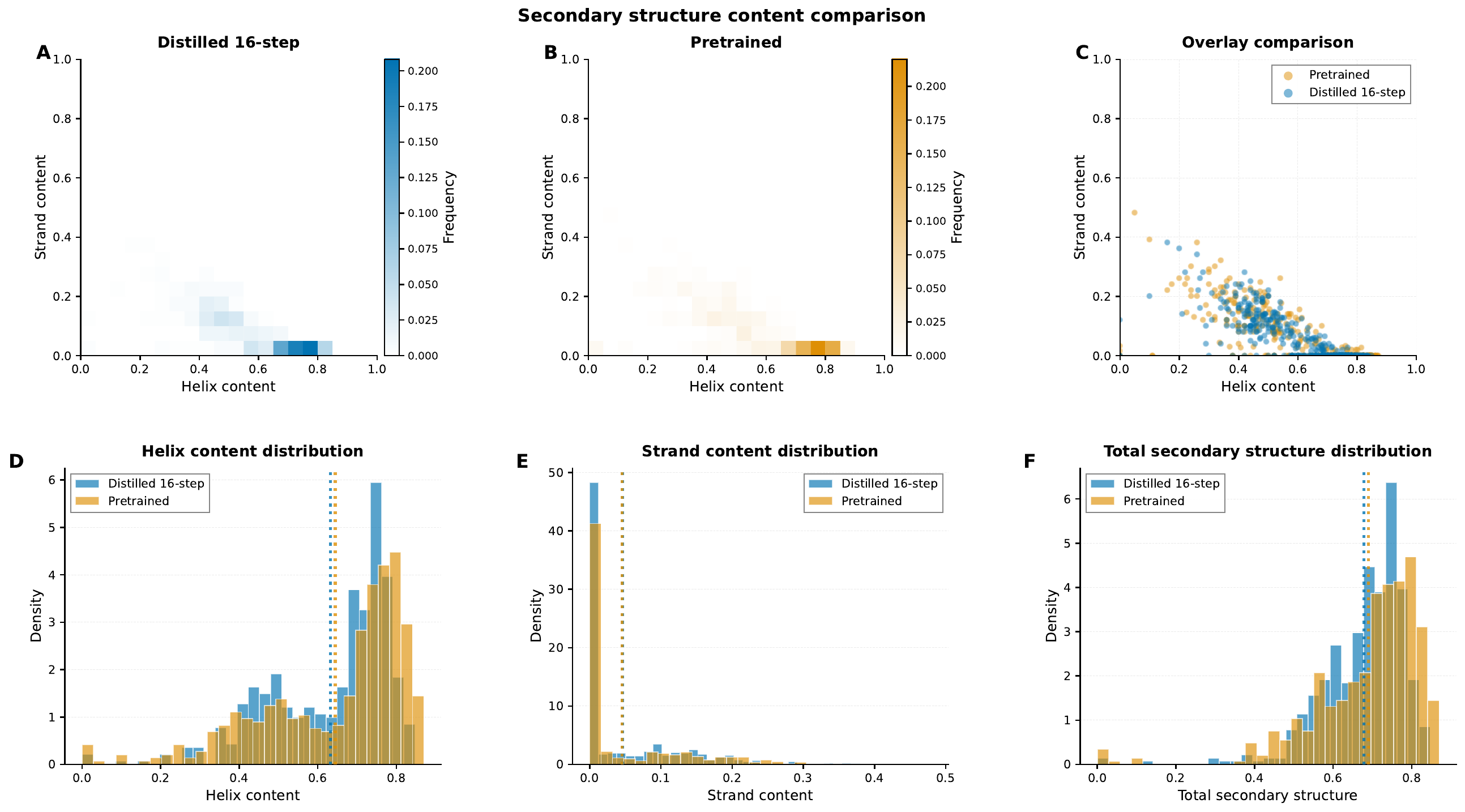}
    \caption{Distribution of secondary structures in the generated protein backbones. Panels A and B show heatmaps for the helix and strand percentages for the 16-step distilled model and the pretrained model respectively. Panel C shows an overlay of each generated backbone. Panels D, E, and F show histograms of the helix percentage, strand percentage, and total secondary structure percentage respectively.}
    \label{fig:SS_distribution}
\end{figure}
We evaluated the structural fidelity of the generated backbones by quantifying their secondary structure content, specifically the percentages of $\alpha$-helices and $\beta$-strands. As illustrated in Fig.~\ref{fig:SS_distribution}, the distributions produced by the distilled model closely align with those of the pretrained baseline. This strongly suggests that SPID faithfully reproduces the structural statistics of the teacher distribution.

\section{Ablation Study and Parameter Settings}
\begin{figure}[ht]
    \vskip 0.2in
    \centering
    \includegraphics[width=\linewidth]{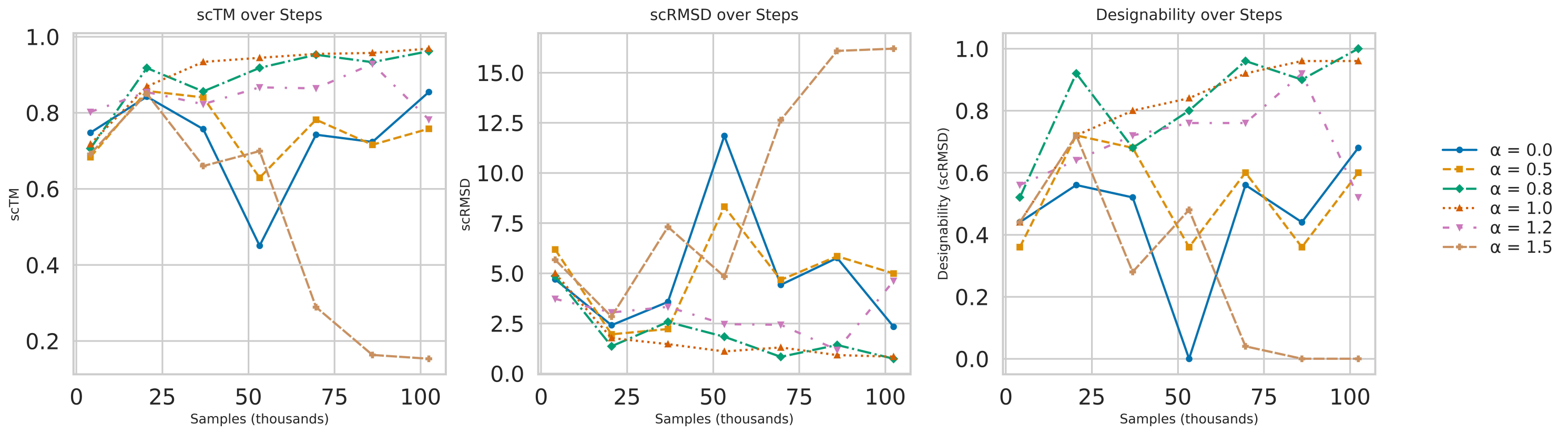}
    \caption{Ablation study of $\alpha$. Each plot illustrates the generation quality, measured in the average scTM, average scRMSD, and scRMSD-based designability within each generated batch against the number of training samples processed during distillation across different $\alpha$ values. The ablation study showcases the effect of $\alpha$ on distillation training and highlights the reason why $\alpha$ is set to 1.0 for subsequent experiments. }
    \label{fig:ablation_alpha}
\end{figure}

\textbf{Impact of $\alpha$.} Similar to the original SiD paper \citep{zhou2024sid}, we conduct an ablation study for the impact of $\alpha$ on the distillation training process. Designability metrics including the average scTM, the average scRMSD, and the designability based on the proportions of structures with scRMSD $<$ 2\r{A} in a batch are plotted in Fig.~\ref{fig:ablation_alpha} as it evolves from 0 to 102.4 thousand samples processed during training. Across the $\alpha$ values of [0, 0.5, 0.8, 1.0, 1.2, 1.5], we see the best designability performance with $\alpha$ being set to 0.8 or 1.0. With other $\alpha$ values, we do not observe a meaningful convergence towards an improved designability. Between the values 0.8 and 1.0, it seems setting $\alpha = 1.0$ results in a slightly smoother and more stable convergence. Hence, we select $\alpha = 1$ for all our experiments.

\textbf{Impact of $\beta_1$}. We follow a similar experimental process as in the SiD paper \citep{zhou2024sid} to investigate the impact of the $\beta_1$ parameter for the Adam optimizer. We compare the performance of setting $\beta_1$ to 0 and 0.9 respectively for the optimizer for the generator $G_\theta$. We did not observe any notable difference in either the convergence speed or the final designability. The $\beta_1$ parameter for the optimizer for the fake score network, $f_\psi$ was kept at 0. As reported in \citet{zhou2024sid}, setting $\beta_1 = 0.9$ often does not result in convergence.

\textbf{Impact of batch size and learning rate.} We initially set the overall batch size to 256 and failed to observe any noticeable improvement in designability. As we increase the batch size to 1024, 2048 and eventually 4096, we observe more and more stable convergence to the level of designability of the pretrained model, with the combination of batch size set to 4096 and learning rate set to 1e-4 and 5e-5 for $f_\psi$ and $G_\theta$ respectively resulting in our best performance. Although it is tempting to conclude that the larger the batch size the better, it is also possible that the unstable performance was due to the learning rate not being optimized for the smaller batch sizes. A more comprehensive grid search might be needed to uncover the optimal setting for the batch size and learning rate. It is also worth noting that the batch size per GPU was set to 2 because the model can be memory-intensive for longer proteins. This results in a large number of gradient accumulation rounds.

\section{Proofs}
As the score function is the key quantity for SiD, we derive it under the flow matching formulation in \proteina. For a pretrained network $f_\phi$ with original data $\xv_d$ and noised sample $\xv_t$ at timestep $t$, the score can be written as:
\begin{align}
    S_\phi(\xv_t) &= \frac{t \vv_t^\phi(\xv_t,t) - \xv_t}{1-t} \\
    &= \frac{t \mathbb{E}(\xv_d - \epsilonv_t | \xv_t) - \xv_t}{1-t} \\
    &= \frac{\mathbb{E}(t\xv_d - t\epsilonv_t - (t\xv_d + (1-t)\epsilonv_t)|\xv_t)}{1-t} \\
    &= -\frac{\mathbb{E}(\epsilonv_t|\xv_t)}{1-t} \\
    &= -\frac{\mathbb{E}(\frac{1}{1-t} (\xv_t-t\xv_d) \given \xv_t)}{1-t} \\
    &= (1-t)^{-2} \mathbb{E}((t\xv_d - \xv_t) \given \xv_t) \\
    &= (1-t)^{-2} (t\mathbb{E}(\xv_d \given \xv_t) - \xv_t) \\
    &= (1-t)^{-2} (t f_\phi(\xv_t,t) - \xv_t)
\end{align}
where $f_\phi(\xv_t, t) = \xv_t + (1-t) v_t^\phi(\xv_t,t)$ and $v_t^\phi$ is the velocity function for $f_\phi$.

The score difference objective in Eq.~\eqref{eq:score_diff} also involves the score of the generator, $\nabla_{\xv_t} \text{ln}~ p_\theta(\xv_t)$. To estimate the generator score, we use another neural network $f_\psi$ and train it using the same training objective as \proteina.
\begin{equation} \label{eq:fake_score_loss_v}
    \text{min}_\psi \mathbb{E}_{q(\xv_t|\xv_g, t) p_\theta(\xv_g)} [\frac{1}{N}\|\vv^\psi_t(\xv_t, t) - (\xv_g - \epsilonv)\|_2^2]
\end{equation}
, where $N$ denotes the number of residues in the desired protein structure.

We can derive the relationship between the network estimation $f_\phi(\xv_t,t)$ and the velocity function $v_t^\phi(\xv_t,t)$:
\begin{align}
    f_\phi(\xv_t,t) &= \mathbb{E}(\xv_d|\xv_t) \\
        &= \mathbb{E}(\frac{1}{t} (\xv_t - (1-t)\epsilonv_t) | \xv_t) \\
        &= \frac{1}{t} (\xv_t - (1-t)\mathbb{E}(\xv_d - \vv_t^\phi(\xv_t,t) | \xv_t)) \\
    \Rightarrow t f_\phi(\xv_t, t) &= \xv_t - (1-t) f_\phi(\xv_t, t) + (1-t) v_t^\phi(\xv_t,t) \\
    \Rightarrow f_\phi(\xv_t, t) &= \xv_t + (1-t) \vv_t^\phi(\xv_t,t) \\
    \Rightarrow \vv_t^\phi(\xv_t,t) &= \frac{1}{1-t} (f_\phi(\xv_t,t)-\xv_t)
\end{align}
and therefore,
\begin{align}
    \vv^\phi_t(\xv_t, t) - (\xv_d - \epsilonv) &= \frac{1}{1-t} (f_\phi(\xv_t,t)-\xv_t) - \xv_d + \epsilonv \\
    &= \frac{1}{1-t}f_\phi(\xv_t,t) - \frac{1}{1-t} (t\xv_d + (1-t)\epsilonv) - \xv_d + \epsilonv \\
    &= \frac{1}{1-t}f_\phi(\xv_t,t) - (\frac{t}{1-t} + 1) \xv_d \\
    &= \frac{1}{1-t} (f_\phi(\xv_t,t)-\xv_d)
\end{align}

Note that the network $f_\psi$ is essentially trained by treating the samples $\xv_g$ produced by the generator $G_\theta$ as the true clean data. Thus, the above relationship can be rewritten for $f_\psi$ as
$\vv^\psi_t(\xv_t, t) - (\xv_g - \epsilonv) = \frac{1}{1-t} (f_\psi(\xv_t,t)-\xv_g)$ and the score of $f_\psi$ is $S_\psi(\xv_t) = (1-t)^{-2}(t\mathbb{E}(\xv_g \given \xv_t)-\xv_t)$.

Hence, Eq.~\eqref{eq:fake_score_loss_v} is equivalent to
\begin{equation}
    \text{min}_\psi \mathbb{E}_{q(\xv_t|\xv_g, t) p_\theta(\xv_g)} [\frac{1}{N(1-t)^2}\|f_\psi(\xv_t, t) - \xv_g\|_2^2]
\end{equation}

We can then approximate the loss in Eq.~\eqref{eq:score_diff} as the score difference between $f_\phi$ and $f_\psi$, and define:
\begin{align}
    \Ls^{(1)}_\theta &= \mathbb{E}_{\xv_t\sim p_\theta(\xv_t)}[\|S_\phi(\xv_t) - S_\psi(\xv_t)\|_2^2] \\
        &= \mathbb{E}_{\xv_t\sim p_\theta(\xv_t)}[\|\frac{t}{(1-t)^2} (f_\phi(\xv_t,t) - f_\psi(\xv_t,t))\|_2^2]
\end{align}

Expanding the $L_2$ norm of the target score difference in Eq.~\eqref{eq:score_diff}, we have:
\begin{align}
    &\mathbb{E}_{\xv_t \sim p_\theta(\xv_t)} [\|S_\phi(\xv_t) - \nabla_{\xv_t} \text{ln}~p_\theta(\xv_t)\|_2^2] \\
    &\approx \mathbb{E}_{\xv_t \sim p_\theta(\xv_t)} [(S_\phi(\xv_t) - S_\psi(\xv_t))^T (S_\phi(\xv_t) - \nabla_{\xv_t} \text{ln}~p_\theta(\xv_t))] \\
    &= \bE_{\xv_t \sim p_\theta(\xv_t)} [(S_\phi(\xv_t) - S_\psi(\xv_t))^T (\frac{1}{(1-t)^2} (t\bE(\xv_d|\xv_t) - \xv_t))] \\&- \bE_{\xv_t \sim p_\theta(\xv_t)} \bE_{\xv_t \sim q(\xv_t\given \xv_g, t)} [(S_\phi(\xv_t) - S_\psi(\xv_t))^T\nabla_{\xv_t} \text{ln}~q(\xv_t\given \xv_g)] \\
    &= \bE_{\xv_t \sim p_\theta(\xv_t)} [(S_\phi(\xv_t) - S_\psi(\xv_t))^T (\frac{1}{(1-t)^2} (t\bE(\xv_d|\xv_t) - \xv_t))] \\&- \bE_{\xv_t \sim p_\theta(\xv_t)} \bE_{\xv_t \sim q(\xv_t\given \xv_g, t)} [(S_\phi(\xv_t) - S_\psi(\xv_t))^T(\frac{1}{(1-t)^2} (t\xv_g - \xv_t))] \\
    &= \frac{t}{(1-t)^2}\bE_{\xv_t \sim p_\theta(\xv_t)} \bE_{\xv_t \sim q(\xv_t\given \xv_g, t)} [(S_\phi(\xv_t) - S_\psi(\xv_t))^T (\bE(\xv_d|\xv_t) - \xv_g)]
\end{align}

Therefore, we define
\begin{align}
    \Ls^{(2)}_\theta &= \frac{t}{(1-t)^2}\bE_{\xv_t \sim p_\theta(\xv_t),\xv_t \sim q(\xv_t\given \xv_g, t)} [(S_\phi(\xv_t) - S_\psi(\xv_t))^T (f_\phi(\xv_t,t) - \xv_g)] \\
    &= \frac{t}{(1-t)^2} \bE_{\xv_t \sim p_\theta(\xv_t),\xv_t \sim q(\xv_t\given \xv_g, t)}[\frac{t}{(1-t)^2}(f_\phi(\xv_t, t) - f_\psi(\xv_t,t))^T (f_\phi(\xv_t,t) - \xv_g)] \\
    &= \frac{t^2}{(1-t)^4} \bE_{\xv_t \sim p_\theta(\xv_t),\xv_t \sim q(\xv_t\given \xv_g, t)}[(f_\phi(\xv_t, t) - f_\psi(\xv_t,t))^T (f_\phi(\xv_t,t) - \xv_g)]
\end{align}

and thus the final SiD loss, $\Ls_\theta = \Ls^{(2)}_\theta - \alpha \Ls^{(1)}_\theta$  as:
\begin{equation}
    \Ls_\theta = \frac{\omega(t)t^2}{(1-t)^4} [(1-\alpha)\|f_\phi(\xv_t,t)-f_\psi(\xv_t, t)\|_2^2 + (f_\phi(\xv_t,t) - f_\psi(\xv_t,t))^T (f_\psi(\xv_t,t) - \xv_g)]
\end{equation}

\section{Evaluation Pipeline}
Our evaluation pipeline largely follows that described in \citet{geffner2025proteina}, with the same metrics for designability, diversity, and novelty, as well as its proposed metrics of FPSD, fS, and fJSD. In the self-consistency pipeline for \textbf{designability}, each generated structure is passed to ProteinMPNN \citep{Dauparas2022ProteinMPNN} to generate 8 candidate sequences. Each of the 8 sequences is fed to ESMFold \citep{Lin2023ESMFold} to predict the corresponding structure. Each ESMFold-predicted structure is then compared with the original generated structure in terms of RMSD. The best RMSD among the 8 predicted structures is reported as the scRMSD. A generated structure is considered designable if the scRMSD is less than 2\r{A}. \textbf{Diversity} is measured by both the average pairwise TM-score \citep{Zhang2004TM} among designable samples and the proportion of designable clusters among all designable samples. \textbf{Novelty} is calculated by first computing the highest TM-score between each generated structure and each structure in the reference dataset, either PDB \citep{berman2000pdb} or AFDB \citep{Varadi2021AFDB}. The average of these maximum TM-scores is reported as the novelty metric. The \textbf{secondary structure content} is calculated based on the P-SEA algorithm \citep{labesse1997psea}. The implementation of the evaluation pipeline, however, is based on both \proteina \citep{geffner2025proteina} and Genie \citep{lin2023genie}. We used the code for the evaluation pipeline provided by Genie to calculate the scRMSD, diversity based on the TM-score, and the secondary structure contents. For the cluster-based diversity and the novelty metrics, we followed the Foldseek commands specified in \proteina. In addition, we use the implementation and the fold class predictor network provided by \proteina for the FPSD, fS, and fJSD metrics. Since the implementation is slightly different, we reran the metrics on the pretrained model instead of copying over the reported results to ensure a fair comparison. For unconditional generation, we generate 100 samples with batch size 10 for protein lengths [50, 100, 150, 200, 250]. We use an Nvidia A6000-48GB GPU and track the total time for generating all samples for timing analyses.

\section{Impact of Number of Sampling Steps}

\begin{figure}
    \vskip 0.2in
    \centering
    \includegraphics[width=\linewidth]{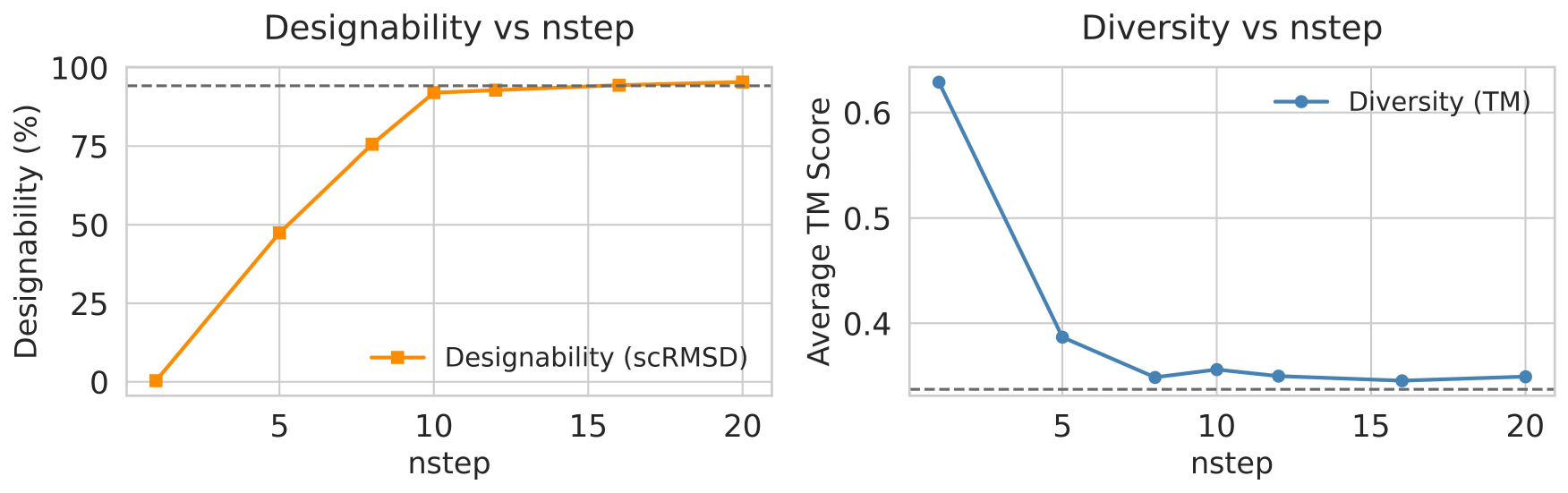}
    \caption{Plots of designability and diversity versus the number of generation steps. On the left, we show the designability as the percentage of generated samples that meet the designable threshold (scRMSD $<$ 2). On the right, we show the diversity as the average pairwise TM-scores between designable samples. The lower the average TM-score is, the more diverse the generated samples are. 16 steps seems enough to match the pretrained model in designability, while being slightly worse in diversity. }
    \label{fig:nstep_discussion}
\end{figure}
Increasing the number of sampling steps strictly improves designability. Excluding $K=1$ which only produced 2 designable samples, cluster diversity is still \emph{not} monotone in step count: it peaks at $K = 8$ (0.76) and is lowest at $K = 16$ (0.57) according to Tab.~\ref{tab:main_result}. TM-score-based diversity is mostly flat for $K \geq 8$ with a slight drop at $K=10$. Designability comparable to the teacher is reached at $K \geq 10$, where every difference lies inside its 95\% interval, and the 16-step generator's designability (94.4\%) is statistically indistinguishable from the teacher's (94.2\%; $p = 0.89$).

\begin{figure}[t]
    \vskip 0.2in
    \centering
    \includegraphics[width=0.6\linewidth]{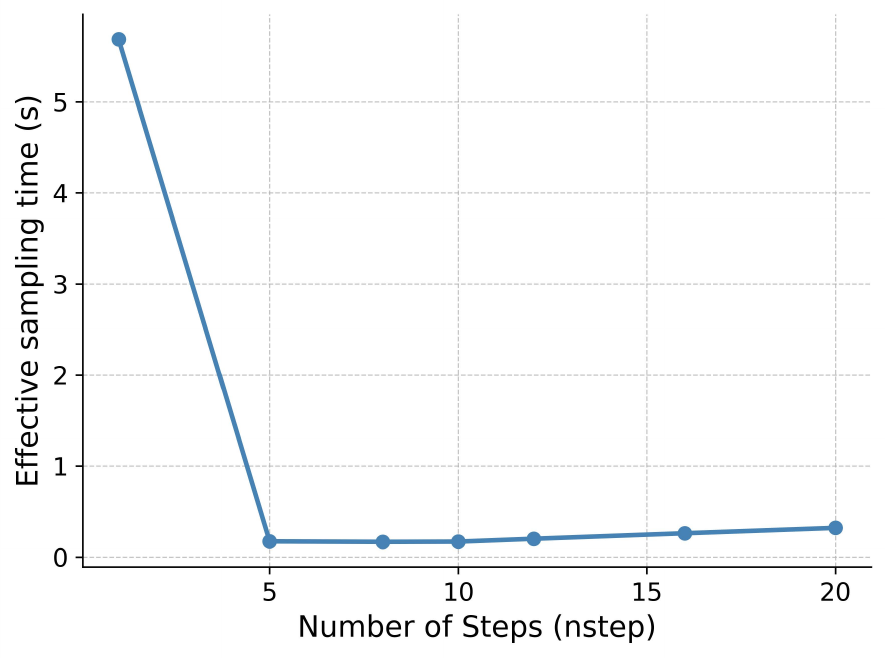}
    \caption{Effective Generation Time against the number of generator steps among distilled generators. 500 structures generated in batches of 10 on an A6000-48GB GPU. The quantity plotted is the \emph{effective backbone-generation} time --- total generation time divided by the number of designable structures. Raw generation time is linear in the step count and is reported in Tab.~\ref{tab:main_result}. }
    \label{fig:time_vs_nstep}
\end{figure}
Although the intuition is that the generation time strictly decreases as the number of sampling steps decrease, it is important to account for the quality of the generated structures in the evaluation of backbone generation time. As shown in Fig.~\ref{fig:time_vs_nstep}, our one-step generator, on average, takes the longest time by far to generate one designable sample compared to other distilled generators. The reason is that the drop in the designability performance outweighs the gain in generation speed for the one-step generator. We argue that the effective generation time more faithfully reflects the speed of generation, compared to simple metrics such as the number of sampling steps or time taken to generate one sample.

\section{Long chain generation}\label{app:long_chain}

\begin{figure}
    \centering
    \includegraphics[width=0.6\linewidth]{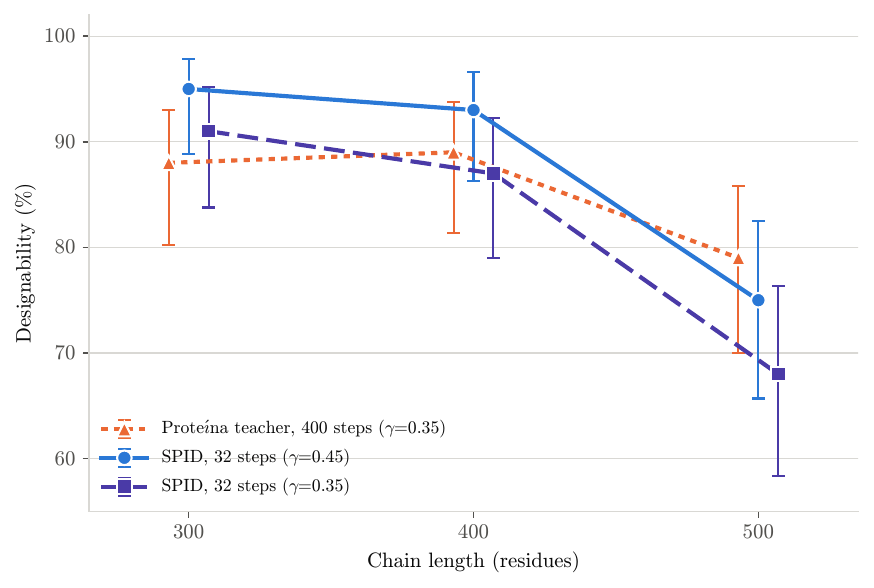}
    \caption{Designability for proteins of lengths in [300, 400, 500]. Orange dotted line represents the designability for the teacher $\mathcal{M}_\text{long}$ rerun. Blue solid line represents our 32-step generator performance at $\gamma=0.45$. Purple dashed line represents our 32-step generator performance at $\gamma=0.35$. Tab.~\ref{tab:long_chain} gives these values numerically, against a baseline measured under our own protocol. }
    \label{fig:long-chain}
\end{figure}

SPID is ahead at $L = 300$ and $400$ and behind by 4 pp at $L = 500$ (75.0 against 79.0), a difference that is not statistically significant ($p = 0.50$), at around 12$\times$ lower effective generation cost and with better cluster diversity and novelty at all three lengths.

In terms of the choice of $\gamma$, Fig.~\ref{fig:long-chain} shows a comparison of $\gamma=0.35$ (same as pretrained teacher) vs $\gamma=0.45$. It seems $\gamma=0.45$ gives slightly better designability scores, even though it might not be statistically significant. Accompanied by the $\gamma$ sweep results in Appendix \ref{app:gamma_frontier}, we decided to report the results for $\gamma=0.45$ in our main text.

We have found that the number of steps is very influential for these longer chains. We had to increase the number of steps to 32 to achieve comparable performance to the pretrained model. This is a genuine scaling limitation rather than an artefact of one checkpoint: at $K = 16$ the long-chain generator does not reach usable designability at these lengths at all. \textbf{The step budget is therefore not a fixed property of the distilled model; it scales with problem size} --- $K = 16$ for lengths 50--256, $K = 32$ for 300--500 --- and the $20\times$ speedup quoted in the main text is specific to the short-chain regime. Designability falls to 75.0\% at $L = 500$ from 95.0\% at $L = 300$, although the measured baseline falls with length too and the gap between them there is not statistically resolvable at $n = 100$.

\textbf{Lengths beyond 512 are out of reach for our training setup.} SiD distillation holds four networks resident plus two optimizer states and backpropagates through the generator, roughly 4$\times$ the memory of inference, which exceeds 48~GB above 512 residues. We therefore do not inherit \proteina\!\!'s stated capability up to 800 residues.

\section{Example of Generated Structures}
\begin{figure}
    \vskip 0.2in
    \centering
    \includegraphics[width=0.9\linewidth]{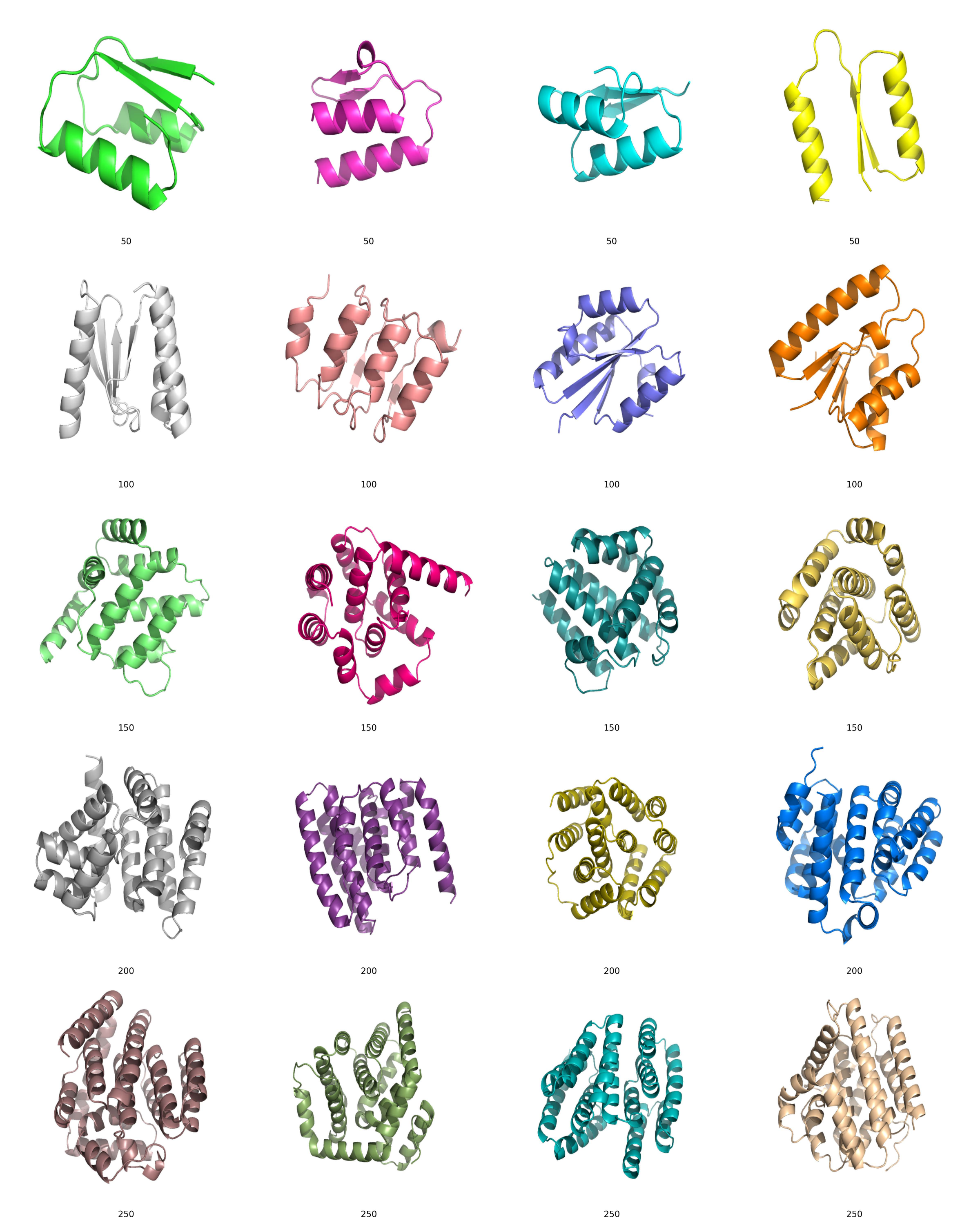}
    \caption{Example of structures generated unconditionally of varying lengths. The lengths are displayed below each structure. All structures shown here are designable. }
    \label{fig:structure_visualization}
\end{figure}

\begin{figure}
    \vskip 0.2in
    \centering
    \includegraphics[width=0.9\linewidth]{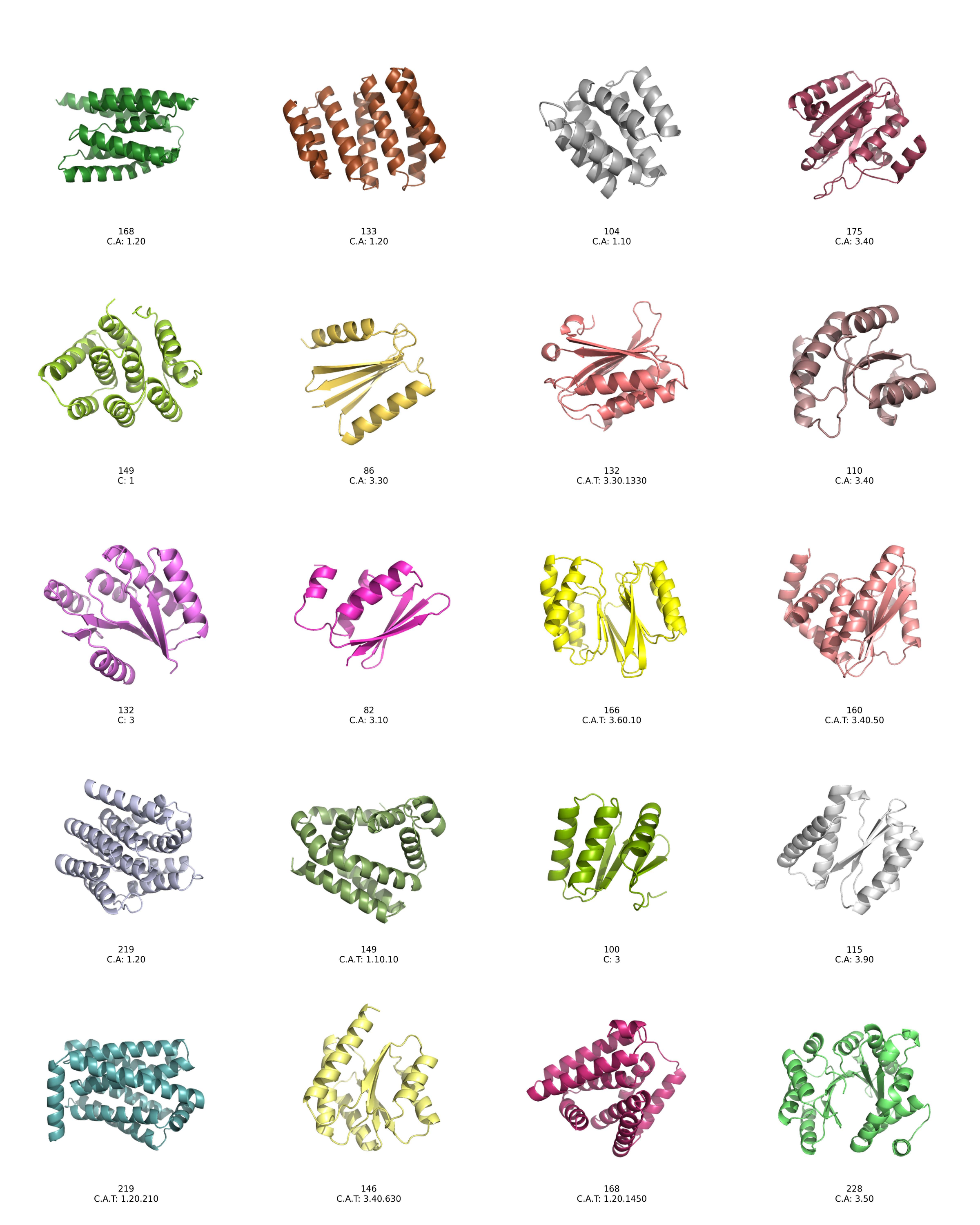}
    \caption{Examples of generated structures conditioned on the fold class as specified by the CATH code. The length and the C.A.T fold class label are denoted below the structures. All structures shown here are designable. }
    \label{fig:conditional_structure_visualization}
\end{figure}

Fig.~\ref{fig:structure_visualization} shows backbones generated unconditionally by our 16-step distilled generator model. Fig.~\ref{fig:conditional_structure_visualization} shows backbones generated while conditioned on fold-class labels and lengths sampled from the empirical distribution of AFDB.
Both sets of examples are randomly selected from the generated structures that are considered designable.

\section{Example of Failure Cases}
\begin{figure}
    \vskip 0.2in
    \centering
    \includegraphics[width=0.9\linewidth]{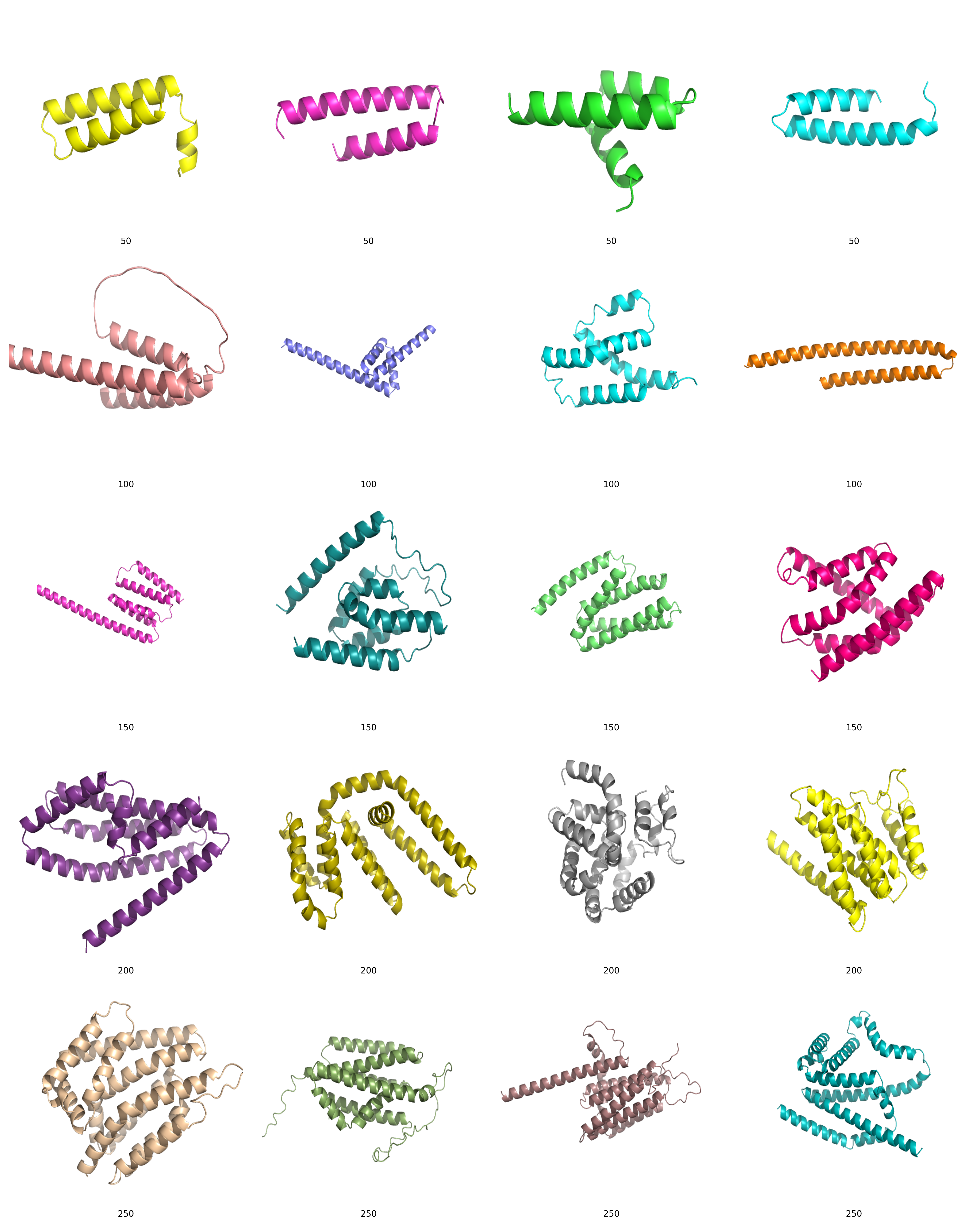}
    \caption{Examples of structures generated unconditionally by our one-step distilled generator that are considered undesignable. The lengths are displayed below each structure. }
    \label{fig:undesignable_example}
\end{figure}
Examining these undesignable structures in Fig.~\ref{fig:undesignable_example} carefully, it is evident that our one-step generator favors alpha helices. No beta sheet has been generated in these samples. Additionally, there exists some exceptionally long alpha helices, such as the ones in the last example of the $N=100$ row and the first sample in the $N=150$ row. Furthermore, the alpha helices tend to appear in 180 degree angles to one another, which is again unusual in stable protein structures. Overall, the one-step generator was not able to model the geometric properties of protein structures.

\section{Additional Ablations and Objective Comparisons}\label{app:ablations}

\subsection{One-step generation: the initial-noise sweep}\label{subsec:one_step}
The one operation resembling noise rescaling available at $K=1$ is to scale the initial Gaussian sample, $\xv_g = G_\theta(t_{init}\cdot 0 + \gamma_0(1-t_{init})\epsilonv)$. We swept $\gamma_0 \in \{0.2, 0.3, 0.45, 0.6, 0.8, 1.0\}$ against the injection timestep $t_{init} \in \{0.2, 0.37, 0.6\}$ at $n=100$ per cell, on \emph{three} generators: the unmodified teacher weights driven through the single step, a one-step generator trained with the Diff-Instruct objective, and a one-step generator trained with the SPID objective. All runs give 0.0\% designability except for the $\gamma_0=1.0, K=1$ setting for the SPID generator which gives 0.4\%. This suggests that the failure is a property of single-step generation rather than of the objective, the training budget, or the injection timestep.

\subsection{Step count against noise scale}
Tab.~\ref{tab:step_gamma_scan} separates the two ingredients along the step axis. Neither substitutes for the other.

\begin{table}[t]
    \centering
    \caption{Designability against step count at both noise scales ($n = 500$). At $K=1$, noise rescaling is not applicable, so we report the same results for both rows. Every other column shows significant improvements in designability when noise rescaling is applied. }
    \label{tab:step_gamma_scan}
    \resizebox{\textwidth}{!}{%
    \begin{tabular}{lccccc}
    \vspace{4pt} \\
    \toprule
    Steps $K$ & 1 & 10 & 16 & 20 & Teacher, 400 \\
    \midrule
    $\gamma = 1$ (no rescaling)   & 0.4 [0.1, 1.4] & 18.2 [15.1, 21.8] & 28.2 [24.4, 32.3] & 37.0 [32.9, 41.3] & 12.4 [9.8, 15.6] \\
    $\gamma = 0.45$               & 0.4 [0.1, 1.4] & 92.0 [89.3, 94.1] & \bf{94.4} [92.0, 96.1] & 95.4 [93.2, 96.9] & 94.2 [91.8, 95.9] \\
    \midrule
    Gain from rescaling (pp)      & --- & $+$73.8 [69.3, 77.6] & $+$66.2 [61.5, 70.3] & $+$58.4 [53.6, 62.8] & $+$81.8 [77.8, 84.9] \\
    \bottomrule
    \end{tabular}
    }
\end{table}

\subsection{The objective comparison across tasks}
Tab.~\ref{tab:objective_by_task} collects the SPID-against-Diff-Instruct comparison on the three tasks this paper addresses. On the fold-class conditional task (Tab.~\ref{tab:conditional_result}) both arms are unconditionally distilled generators conditioned at inference, so no additional training is required. For motif scaffolding tasks, the SPID objective wins on more targets but loses on the total unique successes, so the comparison is inconclusive. 

\begin{table}[t]
    \centering
    \caption{SPID against Diff-Instruct across the three tasks, at matched step count, noise scale, protocol and sample size within each row. Motif scaffolding is measured in unique successes over 24 targets, so its scale is not comparable to the designability columns.}
    \label{tab:objective_by_task}
    \resizebox{\textwidth}{!}{%
    \begin{tabular}{llccll}
    \vspace{4pt} \\
    \toprule
    Task & Metric & SPID & Diff-Instruct & Gap & Diff-Instruct on distributional metrics \\
    \midrule
    Unconditional, $K{=}16$, $n{=}500$ & designability \% & \bf{94.4} & 64.8 & $+$29.6 pp & better (PDB-FPSD 194.7 vs.\ 404.0) \\
    Fold-class conditional, $n{=}900$ & designability \% & \bf{81.4} & 70.4 & $+$11.0 pp, $p = 5{\times}10^{-8}$ & better (PDB-FPSD 202.1 vs.\ 240.0) \\
    Motif scaffolding, 24 targets & unique successes & 1752 & \bf{1835} & $-$83 & --- \\
    \bottomrule
    \end{tabular}
    }
\end{table}

\subsection{Effect of distillation}\label{sec:ablation_contributions}
The importance of the two ingredients of SPID, few-step generation and inference-time noise rescaling, has been established in previous sections. However, the effect of distillation itself has yet to be shown. We can run an exact control: the \emph{unmodified teacher checkpoint} placed inside our few-step sampler, with no distillation of any kind. Tab.~\ref{tab:ablation_2x2} shows that at the same step count ($K=16$), changing the sampler from the \proteina teacher's SDE or ODE solver to our sampling scheme (without noise rescaling) alone can achieve a 25\% designability. Incorporating noise rescaling with $\gamma=0.45$ further improves the designability by 60.4 percentage points to 85.4\%. Finally, distillation adds a further 9 percentage points. This shows that the main advantage comes from the SPID sampler and noise rescaling, although to reach the teacher's designability, distillation training is still needed.

\begin{table}[t]
    \centering
    \caption{Weights $\times$ inference-time rescaling at $K = 16$, $n = 500$, identical evaluation. Rows differ only in the generator weights; columns only in whether $\gamma$ is applied. The bottom row runs the teacher through \emph{its own} sampler at the same step count, for reference. Wilson 95\% intervals in brackets. Distillation is significant at $\gamma = 0.45$ but not at $\gamma = 1$.}
    \label{tab:ablation_2x2}
    \resizebox{\textwidth}{!}{%
    
    \begin{tabular}{lccc}
    \vspace{4pt} \\
    \toprule
    $K = 16$, $n = 500$ & $\gamma = 1$ & $\gamma = 0.45$ & Effect of rescaling \\
    \midrule
    Teacher weights, SPID few-step sampler & 25.0 [21.4, 29.0] & 85.4 [82.0, 88.2] & $+$60.4 pp \\
    Distilled weights (SPID)               & 28.2 [24.4, 32.3] & \bf{94.4 [92.0, 96.1]} & $+$66.2 pp \\
    \midrule
    Effect of distillation                 & $+$3.2 pp ($p = 0.25$) & $+$9.0 pp ($p = 2.3{\times}10^{-6}$) & \\
    \midrule
    \emph{Teacher weights, teacher's own $K{=}16$ sampler} & --- & \emph{0.0 (SDE) / 4.2 (ODE)} & \\
    \bottomrule
    \end{tabular}
    }
\end{table}

\end{document}